\pdfoutput=1

\documentclass[11pt]{article}

\usepackage{authblk}
\usepackage{acl}

\usepackage{times}
\usepackage{latexsym}

\usepackage[T1]{fontenc}

\usepackage[utf8]{inputenc}

\usepackage{microtype}

\usepackage{hyperref}
\usepackage{url}

\usepackage{graphicx}
\usepackage{amsmath}
\usepackage{amssymb}
\usepackage{bbm}
\usepackage{tabularx}
\usepackage{booktabs}
\usepackage{multirow}
\usepackage{array}
\usepackage{mathtools}

\usepackage{xcolor,colortbl}
\usepackage{makecell}

\title{Feature-Adaptive and Data-Scalable In-Context Learning}

\author{Jiahao Li$^1$, Quan Wang$^2$, Licheng Zhang$^1$, Guoqing Jin$^3$, Zhendong Mao$^1$\thanks{\hspace{0.15cm}Corresponding author: Zhendong Mao.}\\ \\
  $^1$University of Science and Technology of China, Hefei, China \\
  $^2$MOE Key Laboratory of Trustworthy Distributed Computing and Service, \authorcr Beijing University of Posts and Telecommunications, Beijing, China \\
  $^3$People’s Daily Online Co., Beijing, China \\
  {\tt \{jiahao66, zlczlc\}@mail.ustc.edu.cn, wangquan@bupt.edu.cn} \\
  {\tt jinguoqing@people.cn, zdmao@ustc.edu.cn}}

\begin{document}
\maketitle
\begin{abstract}
In-context learning (ICL), which promotes inference with several demonstrations, has become a widespread paradigm to stimulate LLM capabilities for downstream tasks. 
Due to context length constraints, it cannot be further improved in spite of more training data, and general features directly from LLMs in ICL are not adaptive to the specific downstream task. 
In this paper, we propose a feature-adaptive and data-scalable in-context learning framework (FADS-ICL), which can leverage task-adaptive features to promote inference on the downstream task, with the supervision of beyond-context samples.
Specifically, it first extracts general features of beyond-context samples via the LLM with ICL input form one by one, and introduces a task-specific modulator to perform feature refinement and prediction after fitting a specific downstream task. 
We conduct extensive experiments on FADS-ICL under varying data settings (4$\sim$128 shots) and LLM scale (0.8$\sim$70B) settings. Experimental results show that FADS-ICL consistently outperforms previous state-of-the-art methods by a significant margin under all settings, verifying the effectiveness and superiority of FADS-ICL. For example, under the 1.5B and 32 shots setting, FADS-ICL can achieve \textbf{+14.3} average accuracy from feature adaptation over vanilla ICL on 10 datasets, with \textbf{+6.2} average accuracy over the previous state-of-the-art method, and the performance can further improve with increasing training data. Code and data are publicly available at \url{https://github.com/jiahaozhenbang/FADS-ICL}.

\end{abstract}

\section{Introduction}

In recent years, increasingly large-scale pre-trained language models (LLMs) have emerged, which leads to a substantial cost for fine-tuning different models for each downstream task. 
Alternatively, in-context learning (ICL) has shown impressive performance on NLP downstream tasks without any modifications to LLMs. 
Specifically, ICL prepends several input-label pairs (demonstrations) to a test sample via a task-specific template and conducts prediction conditioned on this prompt, namely context, for better inference \citep{DBLP:conf/nips/BrownMRSKDNSSAA20}.

Most previous works focus on how to design the best prompt to steer its best performance. 
For example, some works aim to search for the most suitable templates for specific tasks \citep{ DBLP:conf/acl/SorensenRRSRDKF22, DBLP:conf/eacl/PrasadHZB23}, while others focus on selecting the best demonstrations for each test sample \citep{DBLP:conf/acl-deelio/LiuSZDCC22,DBLP:conf/naacl/RubinHB22,DBLP:journals/corr/abs-2307-07164}. 
There are even a few works that explore the order of demonstrations \citep{DBLP:conf/acl/LuBM0S22,DBLP:conf/acl/0003WYK23}.
However, due to the context length constraints of LLMs, these ICL-based methods can only accept a limited number of samples as demonstrations put into the context, even if more labeled samples are available, which severely hinders performance on downstream tasks. 
Thus, what we need is \textbf{data scalability}, that is, the model can accept scalable labeled data to enhance model performance.

\begin{figure}[t!]
\centering
 \includegraphics[width=0.4\textwidth]{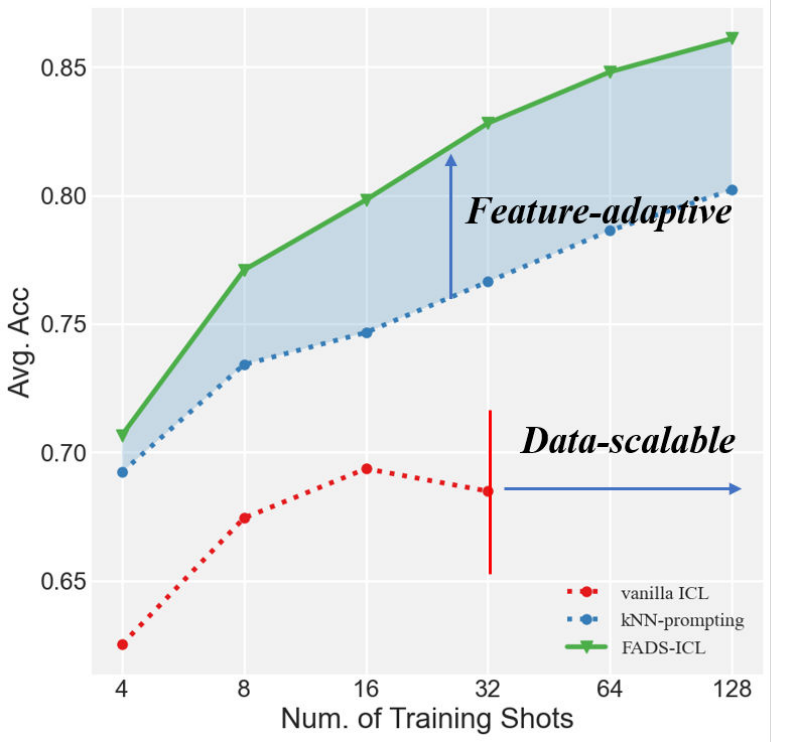}
\caption{The effect of data scalability and feature adaptation on ICL. For data scalability, the performance of vanilla ICL cannot be further improved when exceeding a certain amount of data, but kNN-prompting and FADS-ICL can.  For feature adaptation, FADS-ICL conducts feature refinement for specific tasks, so that it outperforms kNN-prompting using general features by large margins under all data settings. }
\label{fig:intro}
\end{figure}

Recently, a few works have explored exploiting beyond-context samples, similar to KNN-LM \citep{DBLP:conf/iclr/KhandelwalLJZL20}.
Instead of putting all labeled samples into the context at once, they divide the inference process into two parts: sequentially obtaining the feature representation of each sample, and conducting prediction based on the relevance of the feature representation.
Specifically, they compute nearest neighbors for the test sample based on the distance in the feature space and adjust prediction results accordingly. 
For example, kNN-prompt \citep{DBLP:conf/emnlp/ShiMGZ22} performs the final prediction by interpolating the original predicted distribution with the distribution of nearby samples, while kNN-prompting \citep{DBLP:conf/iclr/Xu0MLS023} adopts voting by them directly. 
However, they ignore feature refinement for specific tasks and instead directly use general features obtained from the LLM designed for language modeling, which we refer to as the issue of \textbf{feature adaptation}.
Such non-adaptive feature usage can also severely impair the performance on downstream tasks.
From \autoref{fig:intro}, we can see that under the framework of ICL, data scalability and feature adaptation are both essential to performance on downstream tasks. 

To this end, we proposed a feature-adaptive and data-scalable in-context learning framework (FADS-ICL), which can leverage task-adaptive features to promote inference on the downstream task, with the supervision of beyond-context samples. 
Specifically, FADS-ICL first extracts the general features of each sample by feeding them in the form of vanilla ICL to an LLM in turn. 
Then, these features and corresponding labels are used to supervise the training of a lightweight modulator for feature adaptation in the downstream task. 
Finally, during inference, this modulator can refine the general features to the task-adaptive features for test samples and perform the final prediction based on them.
Here, FADS-ICL solves the issue of data scalability by feeding them in the form of ICL to the LLM one by one to extract general features, which does not need to put all samples into context. 
Besides, FADS-ICL also implements feature adaptation by further refining general features through a modulator for a specific downstream task.

To verify the effectiveness of FADS-ICL, we conduct evaluation experiments on 10 established datasets under varying amounts of labeled samples and LLM scales. 
The main experimental results show that FADS-ICL consistently outperforms state-of-the-art methods by a large margin under all data settings (4$\sim$128 shots, refer to \autoref{fig:intro}) and all LLM scale settings (0.8B$\sim$70B).
Specifically, under the fixed 1.5B LLM and 32 shots, FADS-ICL can achieve \textbf{+14.3} average accuracy from feature adaptation over vanilla ICL on 10 datasets, with \textbf{+6.2} average accuracy over the previous state-of-the-art method, kNN-prompting.
When growing to 128 shots, the improvement over vanilla ICL further increases to \textbf{+17.8}, showing its data scalability.
In short, feature adaptation can improve downstream performance with a fixed amount of available data, while data scalability comes into play as more data becomes available. 

Further, we systematically analyze the impact of three key settings in FADS-ICL, including modulators, selection of general features, and the number of demonstrations, which provides in-depth insights and the corresponding recommended configurations in FADS-ICL.
The analysis experiment on different modulators reveals the importance of a parametric modulator that can promote feature adaptation in FADS-ICL.
Besides, the analysis experiment on the selection of general features shows the hidden states containing rich semantics are more suitable for FADS-ICL than probability distribution. 
Further, our study on the role of demonstrations shows that demonstrations during feature extraction are vital to FADS-ICL while FADS-ICL is robust to the number of demonstrations. 

Our contributions are summarized below:
(1) We propose a feature-adaptive and data-scalable in-context learning framework, which can leverage task-adaptive features to promote inference on the downstream task, with the supervision of beyond-context samples. To our knowledge, this is the first time that feature adaptation has been considered in ICL. 
(2) Extensive experiments are conducted on 10 datasets under varying settings and show that FADS-ICL consistently outperforms previous state-of-the-art methods by a significant margin in almost all settings.
(3) Detailed analytical experiments on key settings provide in-depth insights and recommended configurations in FADS-ICL.

\section{Preliminary}

\subsection{ICL}
ICL is actually a data-enhanced inference framework, which prepends demonstrations to a test sample and conducts prediction conditioned on this prompt for better inference. 
Specifically, assuming a train set $\{(x_i, y_i)\}$, it first needs to randomly sample several demonstrations $D=\{(x_i^D, y_i^D)\}$, of which the number $|D|$ is limited to the context length.
Then, a task-specific template $\mathcal{T}(\cdot)$ (refer to \autoref{appendix:template} for details) is used to wrap up demonstrations along with a test sample $x_{test}$ followed by concatenation to obtain the prompt:
\begin{equation*}
\begin{split}
\mathcal{P}=\mathcal{T}(x_1^D, &\mathcal{V}(y_1^D))\oplus\mathcal{T}(x_2^D, \mathcal{V}(y_2^D))\oplus\cdots\oplus \\
&\mathcal{T}(x_{|D|}^D, \mathcal{V}(y_{|D|}^D))\oplus\mathcal{T}(x_{test},*), 
\end{split}
\label{prompt}
\end{equation*}
where $\mathcal{V}(\cdot)$ is a label verbalizer, mapping each label id to a token in the vocab, {\it e.g.} positive/negative or optimistic/pessimistic in sentiment classification. 
Finally, it takes the prompt as the input of the LLM, and computes the probability distribution by:
\begin{equation}
p(y|x) \propto p_{\theta}(\mathcal{V}(y)|\mathcal{P}), 
\label{ICL}
\end{equation}
with normalization, where $\theta$ denotes the parameters of the LLM. 
Absolutely, ICL can only make use of several demonstrations in context, in spite of the residual in the trainset, namely data non-scalability.

\subsection{KNN-prompting}
To implement data scalability, kNN-prompting tries to obtain probability distributions of residual samples in the same way and vote the test label by $k$ nearest neighbors based on distances between distributions, instead of obtaining the label directly from the distribution. 
Specifically, it first collects probability distributions of residual samples in the trainset by \autoref{ICL}. Then it computes distances between the distribution of the test sample and those of labeled samples based on KL divergence. Finally, $k$ nearest neighbors based on distances are responsible for voting on the test label. 
KNN-prompting implements data scalability by exploiting distributions of the residual in the trainset, but these distributions directly come from the freezing LLM designed for language modeling, that is, ignoring specific adaptation to downstream tasks. 

\begin{figure}[t!]
\centering
 \includegraphics[width=0.45\textwidth]{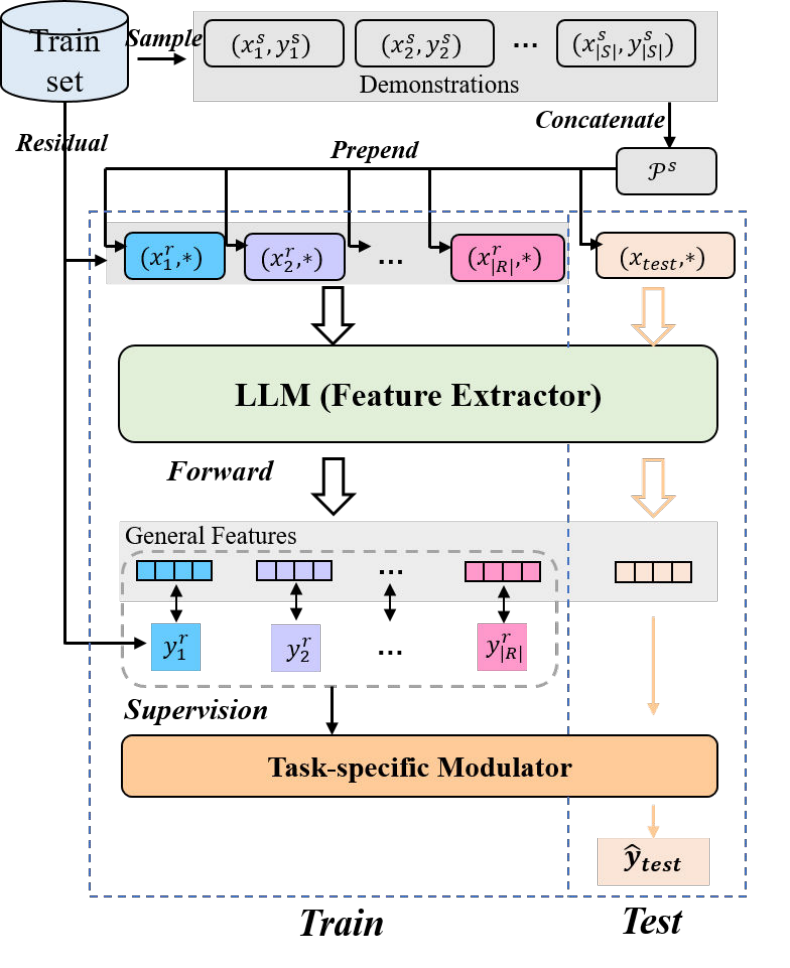}
\caption{The overall framework of FADS-ICL. }
\label{fig:framework}
\end{figure}

\section{FADS-ICL}

For the sake of achieving both feature adaptation and data scalability, we propose FADS-ICL. 
In this section, we elaborate on the architecture of FADS-ICL, including a feature extractor and a lightweight task-specific modulator. 
The overall framework is depicted in \autoref{fig:framework}. 
Note that the train set in FADS-ICL is split to a small demonstration set $S=\{(x_i^s, y_i^s)\}$ and a residual set $R=\{(x_i^r, y_i^r)\}$, with $D=S \cup R$ for fair comparison to ICL.

\subsection{Feature Extractor}

In FADS-ICL, the LLM is actually regarded as a feature extractor with the form of the vanilla ICL, which obtains general features for subsequent usage. 
Specifically, as in ICL, we first randomly choose a few labeled samples as demonstrations\footnote{Note that FADS-ICL is not as sensitive to the number of demonstrations as ICL, of which one can refer to the \autoref{roleofdemons} for detailed analysis, and thus a few demonstrations are enough, {\it e.g.} one sample per class. } $S$, making up the prompt $\mathcal{P}^s$ via a task-specific template. 
Then we feed each template-wrapped sample $x_i$ (labeled sample $x_i^r$ or test sample $x_{test}$) prepended with this identical prompt $\mathcal{P}^s$ into the LLM. 
The general feature of each sample can be obtained by a simple forward pass:
\begin{equation}
h(x_i) = f_{\theta}(\mathcal{T}(x_i,*)|\mathcal{P}^s),
\label{featureextract}
\end{equation}
where we choose the last hidden state from the last layer of the LLM as the common setting\footnote{More settings are discussed in \autoref{featureselection}.}.  
The obtained feature here serves language modeling rather than a specific downstream task, so-called the general feature, so it needs to undergo feature refinement before downstream usage.

\subsection{Task-specific modulator} 
A lightweight modulator is used to refine general features for feature adaptation on a specific downstream task and perform the final prediction. 
As downstream tasks are not visible to the LLM, inevitably there is the problem of task maladaptiveness. 
In other words, general features from the LLM are not well suited for direct downstream usage. 
Except for the heavy fine-tuning on the LLM, an efficient alternative solution is to utilize an extra lightweight module to perform feature adaptation to downstream tasks while keeping the LLM's parameters and generality. 
Specifically, We match the general features extracted from the residual samples in the training set with their corresponding labels, and they are taken as supervisory signals to train a lightweight modulator:
\begin{equation}
\phi=\arg\min_{\phi}\sum_{1\le i\le |R|}\mathcal{L}(g_{\phi}(h(x_i^r)),y_i^r), 
\label{clftrain}
\end{equation}
where the modulator\footnote{Modulator selection is discussed in \autoref{modulatorselection}. } $g_{\phi}(\cdot)$ can be \emph{Logistic Regression}, \emph{Linear SVM}, \emph{MLP}, {\it etc}, and the specific optimization objective $\mathcal{L}$ and algorithm depend on the specific choice of the modulator and possibly the specific task. 
After training, the modulator can be taken as a solver for the specific task.
During inference, the modulator can refine general features for the specific task and map them to the label space without label verbalization in vanilla ICL:
\begin{equation}
p(y|x) = g_{\phi}(h(x)). 
\label{clf}
\end{equation}
In short, FADS-ICL achieves data scalability by the split use of labeled data ({\it i.e.} using a fixed small part of the data as demonstrations and the residual scalable part for supervising the feature adaptation process) and multiple model calls.
Furthermore, FADS-ICL achieves feature adaptation by refining general features into task-specific features via a well-trained task-specific modulator.

\begin{table*}[!t]
\centering
\resizebox{2\columnwidth}{!}{
\begin{tabular}{{ccccccccccc}}
\toprule
&\textbf{SST2}&\textbf{SUBJ}&\textbf{MPQA}&\textbf{AGNews}&\textbf{CB}&\textbf{CR}&\textbf{DBPedia}&\textbf{MR}&\textbf{RTE}&\textbf{TREC}\\
\midrule
\textbf{Avg. Len. of Samples}&53.5&129.1&18.6&239.1&288.2&96.3&281.5&115.6&343.2&50.8\\
\textbf{Num of Classes}&2&2&2&4&3&2&14&2&2&6\\
\textbf{Num. of Shots (TP)}&20 (2\%)&12 (1\%)&39 (0\%)&3 (0\%)&2 (0\%)&14 (4\%)&1 (77\%)&14 (4\%)&4 (0\%)&8 (1\%)\\
\bottomrule
\end{tabular}}
\caption{Data statistics for all used datasets. Num. of Shots (TP) denotes the Maximum number of training shots (per class) allowed by GPT2-xl, {\it i.e.} 1024 tokens of context.  Inside the parentheses are Truncation Probability (TP).} 
\label{table:datasets}
\end{table*}

\begin{table*}[t]
\centering
\small
\resizebox{2\columnwidth}{!}{
\begin{tabular}{{p{0.9cm}l|cccccccccc|c}}
\toprule
\multicolumn{2}{c|}{\textbf{Setting\&Methods}}&\textbf{SST2}&\textbf{SUBJ}&\textbf{MPQA}&\textbf{AGNews}&\textbf{CB}&\textbf{CR}&\textbf{DBPedia}&\textbf{MR}&\textbf{RTE}&\textbf{TREC}&\textbf{AVG}\\
\midrule
\multirow{5}*{$m=4$}&\small\textbf{ICL}&70.4$_{\pm\textrm{6.5}}$&57.3$_{\pm\textrm{10.5}}$&66.8$_{\pm\textrm{8.1}}$&78.2$_{\pm\textrm{6.7}}$&57.9$_{\pm\textrm{9.7}}$&52.0$_{\pm\textrm{3.2}}$&82.0$_{\pm\textrm{2.1}}$&52.0$_{\pm\textrm{3.8}}$&53.0$_{\pm\textrm{1.7}}$&55.4$_{\pm\textrm{3.7}}$&62.5$_{\pm\textrm{5.6}}$\\
&\small\textbf{kNN-prompt}&60.1$_{\pm\textrm9.7}$&76.4$_{\pm\textrm7.8}$&55.7$_{\pm\textrm8.7}$&84.0$_{\pm\textrm5.3}$&42.1$_{\pm\textrm1.0}$&83.3$_{\pm\textrm6.0}$&85.2$_{\pm\textrm2.8}$&78.2$_{\pm\textrm7.7}$&53.5$_{\pm\textrm2.6}$&54.7$_{\pm\textrm3.6}$&67.3$_{\pm\textrm5.5}$\\
&\small\textbf{kNN-prompting}&73.2$_{\pm\textrm12.3}$&70.4$_{\pm\textrm11.0}$&60.9$_{\pm\textrm12.6}$&77.3$_{\pm\textrm10.8}$&55.7$_{\pm\textrm8.3}$&84.9$_{\pm\textrm4.0}$&79.1$_{\pm\textrm1.6}$&82.9$_{\pm\textrm4.5}$&52.3$_{\pm\textrm3.5}$&55.5$_{\pm\textrm10.7}$&69.2$_{\pm\textrm8.0}$\\
&\small\textbf{FADS-ICL}&68.7$_{\pm\textrm6.6}$&83.3$_{\pm\textrm4.3}$&61.6$_{\pm\textrm8.3}$&78.0$_{\pm\textrm6.1}$&52.9$_{\pm\textrm16.7}$&83.2$_{\pm\textrm9.9}$&\underline{91.1$_{\pm\textrm2.8}$}&76.1$_{\pm\textrm10.3}$&53.4$_{\pm\textrm0.3}$&58.4$_{\pm\textrm11.9}$&\textbf{70.7$_{\pm\textrm7.7}$}\\
\midrule
\multirow{5}*{$m=8$}&\small\textbf{ICL}&73.3$_{\pm\textrm{11.6}}$&64.1$_{\pm\textrm{11.3}}$&72.7$_{\pm\textrm{9.0}}$&78.2$_{\pm\textrm{6.7}}$&57.9$_{\pm\textrm{9.7}}$&66.2$_{\pm\textrm{16.7}}$&82.0$_{\pm\textrm{2.1}}$&72.2$_{\pm\textrm{13.9}}$&53.0$_{\pm\textrm{1.7}}$&54.8$_{\pm\textrm{4.2}}$&67.4$_{\pm\textrm{8.7}}$\\
&\small\textbf{kNN-prompt}&68.2$_{\pm\textrm12.2}$&84.4$_{\pm\textrm7.0}$&64.2$_{\pm\textrm7.7}$&83.8$_{\pm\textrm5.1}$&41.1$_{\pm\textrm0.0}$&84.5$_{\pm\textrm9.0}$&84.9$_{\pm\textrm2.8}$&82.5$_{\pm\textrm5.9}$&52.5$_{\pm\textrm0.2}$&53.8$_{\pm\textrm3.9}$&70.0$_{\pm\textrm5.4}$\\
&\small\textbf{kNN-prompting}&81.4$_{\pm\textrm10.8}$&76.0$_{\pm\textrm5.4}$&71.4$_{\pm\textrm7.9}$&84.3$_{\pm\textrm1.2}$&45.7$_{\pm\textrm7.7}$&83.5$_{\pm\textrm2.9}$&89.9$_{\pm\textrm1.4}$&84.5$_{\pm\textrm4.0}$&53.6$_{\pm\textrm6.1}$&63.7$_{\pm\textrm9.4}$&73.4$_{\pm\textrm5.7}$\\
&\small\textbf{FADS-ICL}&82.2$_{\pm\textrm6.9}$&85.4$_{\pm\textrm4.7}$&74.0$_{\pm\textrm2.4}$&84.2$_{\pm\textrm2.6}$&57.1$_{\pm\textrm4.2}$&88.0$_{\pm\textrm4.7}$&\underline{95.3$_{\pm\textrm1.9}$}&81.3$_{\pm\textrm7.6}$&51.1$_{\pm\textrm2.7}$&\underline{72.2$_{\pm\textrm6.8}$}&\textbf{77.1$_{\pm\textrm4.4}$}\\
\midrule
\multirow{5}*{$m=16$}&\small\textbf{ICL}&81.3$_{\pm\textrm{5.4}}$&64.1$_{\pm\textrm{11.3}}$&83.9$_{\pm\textrm{1.4}}$&78.2$_{\pm\textrm{6.7}}$&57.9$_{\pm\textrm{9.7}}$&66.2$_{\pm\textrm{16.7}}$&82.0$_{\pm\textrm{2.1}}$&72.2$_{\pm\textrm{13.9}}$&53.0$_{\pm\textrm{1.7}}$&54.8$_{\pm\textrm{4.2}}$&69.4$_{\pm\textrm{7.3}}$\\
&\small\textbf{kNN-prompt}&79.2$_{\pm\textrm13.6}$&85.9$_{\pm\textrm5.0}$&75.2$_{\pm\textrm6.7}$&83.4$_{\pm\textrm6.3}$&42.1$_{\pm\textrm1.0}$&82.3$_{\pm\textrm7.3}$&85.8$_{\pm\textrm2.1}$&79.7$_{\pm\textrm7.6}$&52.4$_{\pm\textrm0.2}$&54.7$_{\pm\textrm4.1}$&72.1$_{\pm\textrm5.4}$\\
&\small\textbf{kNN-prompting}&78.7$_{\pm\textrm15.8}$&80.0$_{\pm\textrm2.9}$&73.9$_{\pm\textrm7.5}$&85.4$_{\pm\textrm1.9}$&48.9$_{\pm\textrm5.7}$&82.7$_{\pm\textrm4.4}$&93.5$_{\pm\textrm0.4}$&83.6$_{\pm\textrm3.7}$&52.3$_{\pm\textrm3.2}$&67.7$_{\pm\textrm4.7}$&74.7$_{\pm\textrm5.0}$\\
&\small\textbf{FADS-ICL}&86.6$_{\pm\textrm7.8}$&89.1$_{\pm\textrm2.3}$&79.2$_{\pm\textrm1.7}$&86.5$_{\pm\textrm2.1}$&63.9$_{\pm\textrm3.4}$&88.7$_{\pm\textrm3.4}$&\underline{97.2$_{\pm\textrm0.6}$}&82.5$_{\pm\textrm3.4}$&48.8$_{\pm\textrm3.3}$&\underline{75.7$_{\pm\textrm6.0}$}&\textbf{79.8$_{\pm\textrm3.4}$}\\
\midrule
\multirow{5}*{$m=32$}&\small\textbf{ICL}&81.3$_{\pm\textrm{5.4}}$&64.1$_{\pm\textrm{11.3}}$&75.2$_{\pm\textrm{8.8}}$&78.2$_{\pm\textrm{6.7}}$&57.9$_{\pm\textrm{9.7}}$&66.2$_{\pm\textrm{16.7}}$&82.0$_{\pm\textrm{2.1}}$&72.2$_{\pm\textrm{13.9}}$&53.0$_{\pm\textrm{1.7}}$&54.8$_{\pm\textrm{4.2}}$&68.5$_{\pm\textrm{8.1}}$\\
&\small\textbf{kNN-prompt}&80.9$_{\pm\textrm12.9}$&88.0$_{\pm\textrm3.4}$&72.2$_{\pm\textrm6.0}$&83.6$_{\pm\textrm5.0}$&41.8$_{\pm\textrm1.6}$&85.2$_{\pm\textrm4.0}$&85.7$_{\pm\textrm2.6}$&84.5$_{\pm\textrm5.2}$&52.4$_{\pm\textrm0.2}$&55.0$_{\pm\textrm3.9}$&72.9$_{\pm\textrm4.5}$\\
&\small\textbf{kNN-prompting}&83.9$_{\pm\textrm7.6}$&83.0$_{\pm\textrm1.6}$&71.8$_{\pm\textrm7.9}$&85.3$_{\pm\textrm1.5}$&55.0$_{\pm\textrm6.4}$&86.4$_{\pm\textrm3.3}$&94.5$_{\pm\textrm0.5}$&84.5$_{\pm\textrm1.2}$&51.5$_{\pm\textrm5.8}$&70.6$_{\pm\textrm2.9}$&76.6$_{\pm\textrm3.9}$\\
&\small\textbf{FADS-ICL}&88.0$_{\pm\textrm4.6}$&\underline{90.1$_{\pm\textrm2.2}$}&\underline{81.4$_{\pm\textrm3.0}$}&84.9$_{\pm\textrm1.4}$&\underline{72.1$_{\pm\textrm2.0}$}&\underline{92.3$_{\pm\textrm0.7}$}&\underline{97.9$_{\pm\textrm0.9}$}&85.9$_{\pm\textrm2.8}$&55.5$_{\pm\textrm2.9}$&\underline{79.9$_{\pm\textrm4.7}$}&\textbf{82.8$_{\pm\textrm2.5}$}\\
\midrule
\multirow{5}*{$m=64$}&\small\textbf{ICL}&81.3$_{\pm\textrm{5.4}}$&64.1$_{\pm\textrm{11.3}}$&75.2$_{\pm\textrm{8.8}}$&78.2$_{\pm\textrm{6.7}}$&57.9$_{\pm\textrm{9.7}}$&66.2$_{\pm\textrm{16.7}}$&82.0$_{\pm\textrm{2.1}}$&72.2$_{\pm\textrm{13.9}}$&53.0$_{\pm\textrm{1.7}}$&54.8$_{\pm\textrm{4.2}}$&68.5$_{\pm\textrm{8.1}}$\\
&\small\textbf{kNN-prompt}&83.0$_{\pm\textrm6.8}$&87.3$_{\pm\textrm4.8}$&73.4$_{\pm\textrm8.2}$&83.8$_{\pm\textrm5.7}$&41.4$_{\pm\textrm0.8}$&86.5$_{\pm\textrm3.0}$&86.4$_{\pm\textrm2.1}$&85.2$_{\pm\textrm4.0}$&52.3$_{\pm\textrm0.0}$&54.7$_{\pm\textrm4.4}$&73.4$_{\pm\textrm4.0}$\\
&\small\textbf{kNN-prompting}&87.2$_{\pm\textrm2.8}$&82.5$_{\pm\textrm1.2}$&82.9$_{\pm\textrm4.1}$&86.6$_{\pm\textrm2.1}$&61.4$_{\pm\textrm4.8}$&89.1$_{\pm\textrm1.7}$&95.2$_{\pm\textrm1.2}$&84.3$_{\pm\textrm1.9}$&49.1$_{\pm\textrm3.3}$&67.9$_{\pm\textrm5.5}$&78.6$_{\pm\textrm2.9}$\\
&\small\textbf{FADS-ICL}&88.2$_{\pm\textrm1.8}$&90.5$_{\pm\textrm1.7}$&83.6$_{\pm\textrm1.5}$&87.8$_{\pm\textrm1.0}$&\underline{78.6$_{\pm\textrm3.6}$}&\underline{91.2$_{\pm\textrm0.8}$}&\underline{98.1$_{\pm\textrm0.7}$}&87.0$_{\pm\textrm1.3}$&\underline{58.7$_{\pm\textrm3.7}$}&\underline{84.4$_{\pm\textrm4.0}$}&\textbf{84.8$_{\pm\textrm2.0}$}\\
\midrule
\multirow{5}*{$m=128$}&\small\textbf{ICL}&81.3$_{\pm\textrm{5.4}}$&64.1$_{\pm\textrm{11.3}}$&75.2$_{\pm\textrm{8.8}}$&78.2$_{\pm\textrm{6.7}}$&57.9$_{\pm\textrm{9.7}}$&66.2$_{\pm\textrm{16.7}}$&82.0$_{\pm\textrm{2.1}}$&72.2$_{\pm\textrm{13.9}}$&53.0$_{\pm\textrm{1.7}}$&54.8$_{\pm\textrm{4.2}}$&68.5$_{\pm\textrm{8.1}}$\\
&\small\textbf{kNN-prompt}&86.1$_{\pm\textrm3.9}$&87.5$_{\pm\textrm3.3}$&74.2$_{\pm\textrm4.0}$&84.0$_{\pm\textrm5.3}$&42.5$_{\pm\textrm1.5}$&87.0$_{\pm\textrm3.5}$&86.4$_{\pm\textrm2.2}$&87.3$_{\pm\textrm2.1}$&52.3$_{\pm\textrm0.0}$&55.3$_{\pm\textrm4.1}$&74.3$_{\pm\textrm3.0}$\\
&\small\textbf{kNN-prompting}&86.0$_{\pm\textrm2.8}$&83.9$_{\pm\textrm2.2}$&82.1$_{\pm\textrm1.6}$&87.8$_{\pm\textrm1.5}$&65.0$_{\pm\textrm3.0}$&88.8$_{\pm\textrm2.2}$&96.8$_{\pm\textrm1.0}$&86.2$_{\pm\textrm0.8}$&51.4$_{\pm\textrm1.8}$&74.0$_{\pm\textrm3.2}$&80.2$_{\pm\textrm2.0}$\\
&\small\textbf{FADS-ICL}&88.9$_{\pm\textrm0.8}$&\underline{92.1$_{\pm\textrm0.9}$}&\underline{84.8$_{\pm\textrm1.2}$}&88.4$_{\pm\textrm1.1}$&\underline{83.2$_{\pm\textrm4.5}$}&90.5$_{\pm\textrm1.9}$&\underline{98.6$_{\pm\textrm0.7}$}&86.1$_{\pm\textrm1.0}$&57.8$_{\pm\textrm4.1}$&\underline{90.5$_{\pm\textrm1.6}$}&\textbf{86.1$_{\pm\textrm1.8}$}\\
\bottomrule
\end{tabular}}
\caption{Results under varying data settings. We re-implement kNN-prompt, while kNN-prompting is reproduced using the released code\footnotemark. Here, the LLM scale is 1.5B. \underline{Underline} indicates leading all baselines with $p<0.05$. }
\label{table:mainresultsacrossshot}
\end{table*}
\footnotetext{\url{https://github.com/BenfengXu/KNNPrompting}}

\section{Experiments}

\subsection{Experimental Settings}

\paragraph{Datasets} Following kNN-prompting \citep{DBLP:conf/iclr/KhandelwalLJZL20}, we established our experiments on 10 publicly available datasets, respectively four sentiment classification datasets: SST2 \citep{DBLP:conf/emnlp/SocherPWCMNP13}, MPQA \citep{DBLP:journals/lre/WiebeWC05}, CR \citep{DBLP:conf/kdd/HuL04}, and MR \citep{DBLP:conf/acl/PangL05}, two natural language inference datasets: CB \citep{de2019commitmentbank} and RTE \citep{DBLP:conf/mlcw/DaganGM05}, three topic classification datasets: AGNews \citep{DBLP:conf/nips/ZhangZL15}, DBPedia \citep{DBLP:conf/nips/ZhangZL15}, and TREC \citep{DBLP:conf/sigir/VoorheesT00}, one subjectivity judgment dataset: SUBJ \citep{DBLP:conf/acl/PangL04}. 
The data statistics for all datasets are shown in \autoref{table:datasets}. 
\paragraph{LLMs} 
To verify the effectiveness of FADS-ICL on different LLM scales, we consider a wide range of LLM scales, including GPT2 series (0.8B and 1.5B), Llama-1 series (7B, 13B, and 30B), and Llama-2 series (7B, 13B and 70B)\footnote{Bitsandbytes for 8-bit quantification of Llama-2 (70B). }. 
GPT2-xl (1.5B) is used for most subsequent experiments unless explicitly stated. 

\paragraph{Other Settings or Details} For data setting, we consider varying training shots $m=\{4, 8, 16, 32, 64, 128\}$, which means the number of samples per class. 
We implement all modulators by scikit-learn \citep{sklearn_api}, and conduct both the training and inference processes of modulators on the CPU. Note that even so, the total running time of the modulator is within a second \footnote{Refer to \autoref{subsec:mainresult} for detailed overhead comparison. } and it is negligible compared to the forward pass of the LLM.  
Besides, without special instructions, the lightweight modulator we choose is \emph{Logistic Regression} with the last hidden states as general features, and the number of 
 demonstrations used in FADS-ICL is one sample per class. 

\paragraph{Evaluation} Accuracy is the direct metric for all datasets. We run each experiment with five different seeds, and average accuracy and standard deviation are reported. 

\paragraph{Baselines} 
In addition to vanilla ICL, we also consider stronger baselines for fair comparison. 
kNN-prompt and kNN-prompting can both exploit beyond-context samples and improve the inference performance by interpolation or voting by k nearest neighbors. 
Besides, since sometimes ICL can not put all labeled samples as demonstrations into context, some methods have emerged to improve ICL by selecting the most appropriate demonstrations for each test sample during inference.
We put the detailed comparison with demonstration selection methods in \autoref{APP:comparisontoselection}.

\begin{figure*}[t!]
\centering
\includegraphics[width=\textwidth]{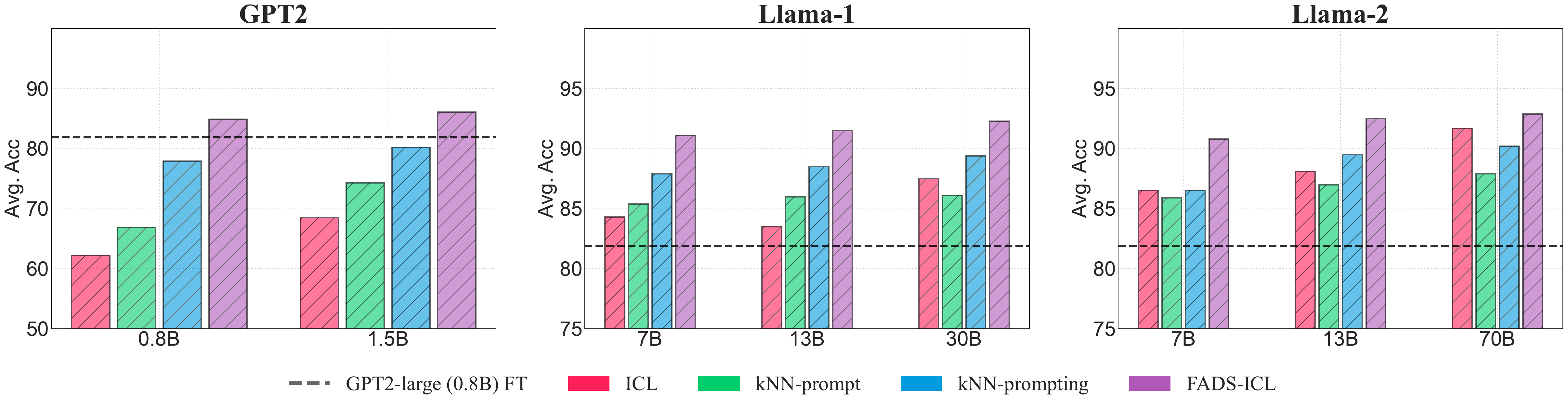}
\caption{Results across LLM scales with 128-shots. }
\label{fig:LLMscale}
\end{figure*}

\begin{figure}[t!]
\centering
\includegraphics[width=0.4\textwidth]{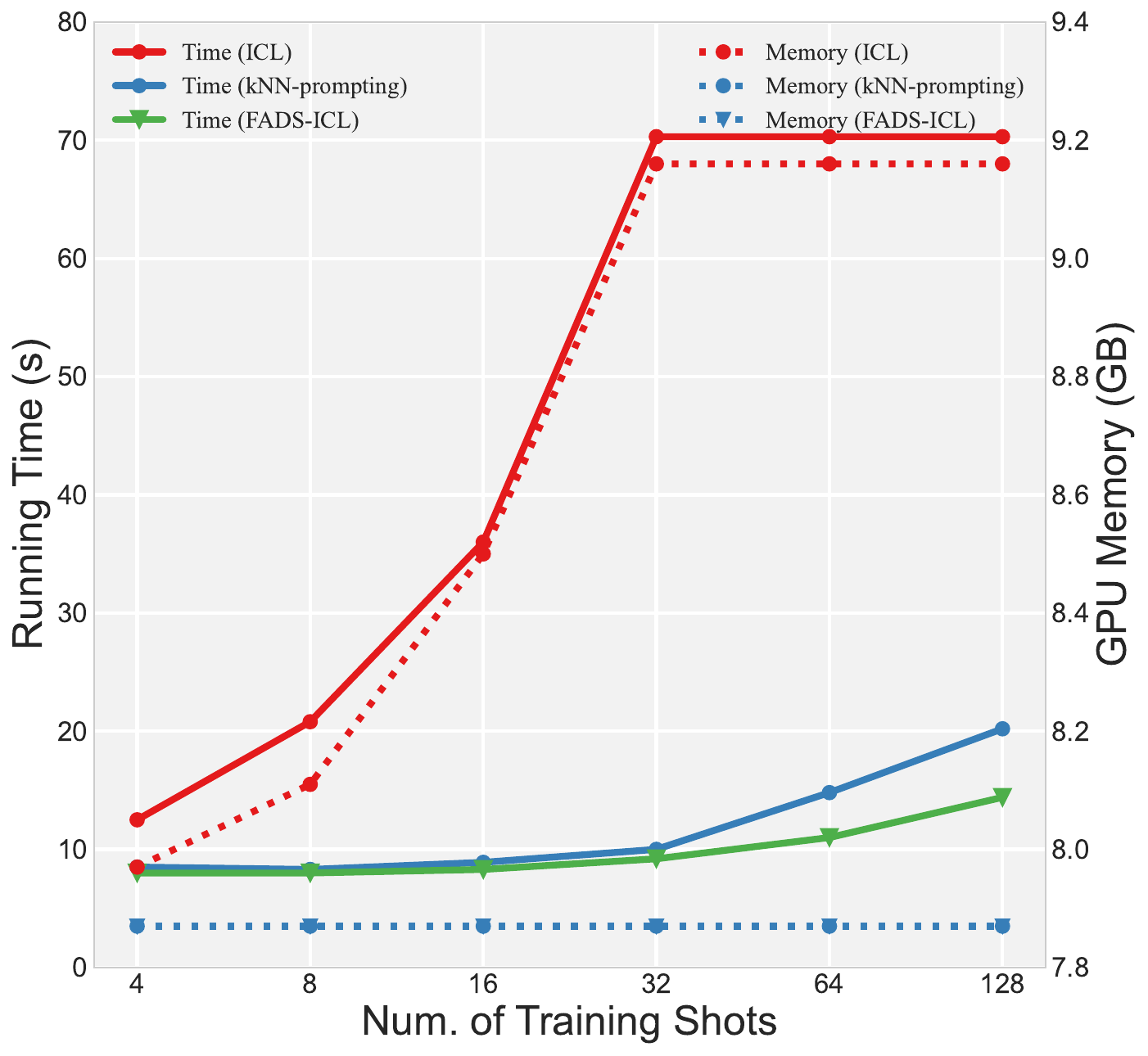}
\caption{The comparison for computational and memory overhead on the MPQA dataset (256 test samples).}
\label{fig:overhead}
\end{figure}

\subsection{Main Results}
\label{subsec:mainresult}
\autoref{table:mainresultsacrossshot} and \autoref{fig:LLMscale} show the results across varying data settings and LLM scale settings\footnote{Full results are listed in \autoref{APP:Performance Across LLM Scales}.} respectively. Experimental results show that FADS-ICL consistently outperforms all baselines under all settings by a significant margin, verifying its effectiveness and superiority on downstream tasks.

\paragraph{For Varying Data Settings} From \autoref{table:mainresultsacrossshot}, the performance of all methods improves significantly as available samples grow. 
However, due to data non-scalability vanilla ICL stops improving and keeps 68.5 average accuracy on 10 datasets after training shots reach 32. 
FADS-ICL can achieve \textbf{+14.3} improvement over ICL with 32 shots by feature adaptation, and can further achieve \textbf{+17.6} improvement with 128 shots, showing data scalability. 
Besides, compared to the previous state-of-the-art method, kNN-prompting, FADS-ICL can achieve the improvement of \textbf{+6.2} average accuracy with 32 shots, and this large margin roughly keeps as available data scales up.
From the perspective of stability,  KNN-based methods and FADS-ICL can both gradually lower the standard deviation of performance along with increasing data, while FADS-ICL can consistently obtain better stability than others under fixed data settings. 

\paragraph{For Varying LLM Scale Settings} From \autoref{fig:LLMscale}, we can see the performance of all methods achieve consistent improvement roughly as the LLM scale increases along with the capability of the LLM itself, and FADS-ICL still consistently performs much better than all other baselines. 
Besides, with the increasing context length (the context length of GPT-2, Llama-1, and Llama-2 series is respectively 1k, 2k, and 4k.), the number of demonstrations used in ICL will increase
proportionally. Thus, the performance of ICL has been significantly improved and is even comparable to kNN-prompting on the Llama-2 series, which does not benefit from increasing context length, but there is still a considerable gap from FADS-ICL which can refine general features into task-specific features.
Surprisingly, under the 0.8B setting, FADS-ICL obtains the best 84.9 average accuracy, even exceeding 81.9 for the entire LLM fine-tuning, which requires fitting the parameters that are 5 orders of magnitude larger than FADS-ICL. 
It reveals that although fine-tuning is always the best way to adapt to downstream tasks, FADS-ICL effectively fitting extremely few designed parameters, can achieve the best performance at minimal cost in few-shot scenarios. 

\paragraph{Overhead Comparison} \autoref{fig:overhead} shows the comparison for computational and memory overhead of FADS-ICL and baselines. We can see that FADS-ICL is significantly lower than ICL in both running time and GPU memory, especially with more available samples. This is because ICL puts as many demonstrations into context as possible and thus the computational and memory overhead will increase nearly quadratically to compute cross-attention within the LLM as available samples increase, and correspondingly, the input length increases. In comparison, FADS-ICL takes one sample per class as a demonstration all the time while it needs several forward passes of the LLM equal to the number of available samples, which causes the computational overhead to increase linearly and the memory overhead unchanged. Considering both points, the overall computational and memory overhead of FADS-ICL are even smaller than those of vanilla ICL. Note again that the overall overhead (including fitting and inference) of the newly introduced modulator in FADS-ICL is about 0.2s on the CPU, which is negligible compared to the LLM overhead. Besides, kNN-prompting takes similar overheads to FADS-ICL due to the same input form, while it takes a little more running time to compute distances between labeled examples and each test sample rather than directly perform prediction based on task-specific features in FADS-ICL.

\begin{figure*}[t!]
\centering
\includegraphics[width=\textwidth]{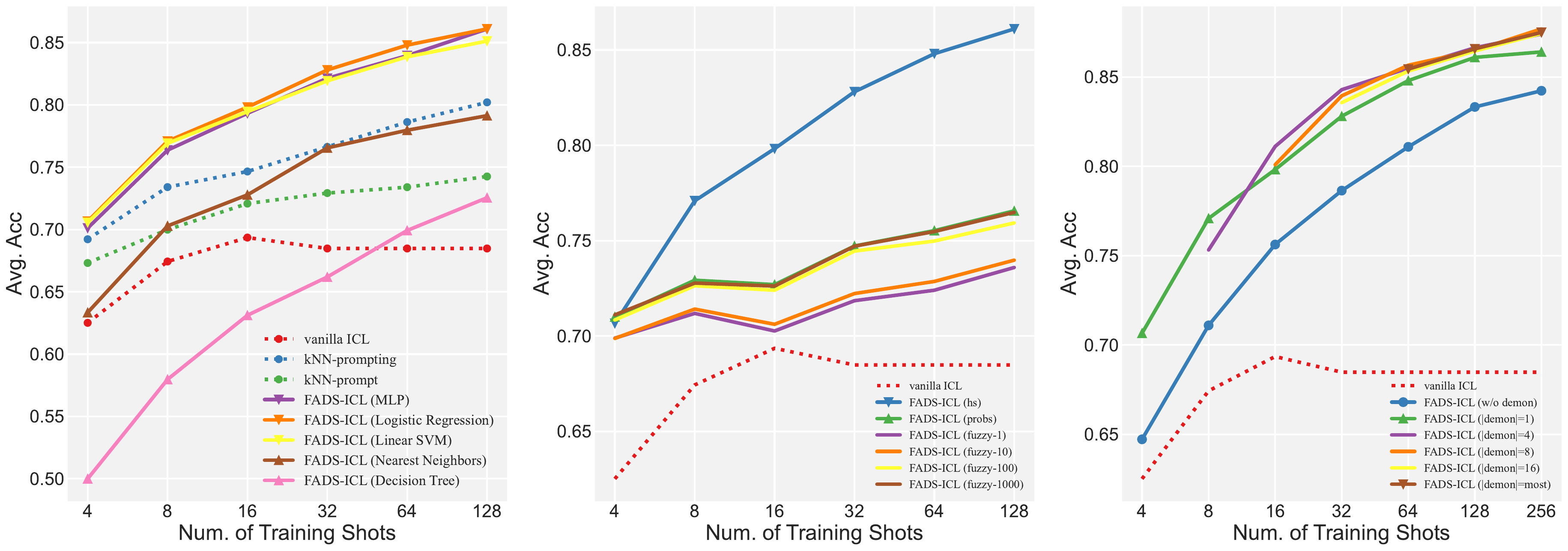}
\caption{\textbf{Left (a):} The effect of different modulators in FADS-ICL. \textbf{Middle (b):} The effect of different features as the general features in FADS-ICL. \textbf{Right (c):} The role of demonstrations in FADS-ICL.  }
\label{fig:Ablation}
\end{figure*}

\section{Analysis}

In this section, we first systematically analyze the impacts of three key settings in FADS-ICL including modulators, selection of general features, and the number of demonstrations, which provides in-depth insights and the corresponding recommended configurations in FADS-ICL.
Visualization for feature adaptation is presented at last. 

\subsection{Effect of Modulators}
\label{modulatorselection}

The modulator contributes to refining features for downstream tasks and performing the final prediction, of which the choice is quite important to FADS-ICL. In this section, we conduct a comparative experiment on parametric modulators and non-parametric modulators, to explore the effect of different modulators in FADS-ICL.

\paragraph{Setup} For the choice of modulators in FADS-ICL, we consider two groups: parametric modulators and non-parametric modulators. The parametric modulators include \emph{MLP}, \emph{Logistic Regression}, and \emph{Linear SVM}, while the non-parametric modulators involve \emph{Nearest Neighbors} and \emph{Decision Tree}. Almost all hyper-parameters in modulators are set as default values in scikit-learn.

\paragraph{Result} From \autoref{fig:Ablation}(a), we can see the parametric modulator group consistently outperforms all other baselines and the non-parametric group by large margins under all data settings, and there are only negligible performance gaps in the parametric modulator group. 
This shows the superiority of the parametric modulator in FADS-ICL, which can further refine general features from the LLM and adapt to downstream tasks better. 
For non-parametric modulators, \emph{Nearest Neighbors} is more superior and stable than \emph{Decision Tree}. 

\paragraph{Analysis and Recommendation} First, we just need to consider parametric modulators, as all three of them exhibit strong performance with only minor differences. Next, considering the number of parameters, \emph{Logistic Regression} and \emph{Linear SVM} are preferred. 
Further, \emph{Logistic Regression} is more simple and efficient than \emph{Linear SVM}, which needs more training time. 
In conclusion, the modulator we recommend is \emph{Logistic Regression}. 


\subsection{Effect of Feature Choice}
\label{featureselection}

In this subsection, we explore the effect of different choices of general features directly extracted from the LLM in FADS-ICL. 

\paragraph{Setup} For the choice of the general feature in FADS-ICL, we consider two kinds: the last hidden state and probability distribution on vocab. Considering the sparsity of probability distribution, we explore extra settings for probability distribution, {\it i.e.} fuzzy-k distribution, $k \in \{1, 10, 100, 1000\}$. 
Specifically, we choose probabilities of the top k most similar tokens to each verbalized label as features used in FADS-ICL. For example, fuzzy-1 probability denotes that only the probabilities of verbalized labels are involved in FADS-ICL.

\paragraph{Result}
From \autoref{fig:Ablation}(b), we can see the FADS-ICL with the last hidden states as general features consistently outperforms the other group of probability distribution when the training shots surpass 4, and the gap becomes increasingly significant with increasing data. 
Besides, we observe another consistent phenomenon within the probability distribution group: with more dimensions remaining, performance consistently improves. 
This suggests that in FADS-ICL, selection for feature dimensions is not necessary, and even dimensions with low information content can be effectively utilized.


\paragraph{Analysis and Recommendation}

In knowledge distillation, \citet{DBLP:conf/emnlp/JiaoYSJCL0L20} shows that hidden states contain richer knowledge than probability distributions in general scenarios, while probability distributions can contribute to specific tasks as soft labels. 
For FADS-ICL, what we need are general features to prepare for subsequent refinement.  
Therefore, if there are enough labeled samples as supervision, extracting task-adaptive features from hidden states tends to perform much better. 
Additionally, with few available samples, {\it e.g. $m=4$}, choosing probability distributions with simple surface semantic information is slightly more suitable for efficiently fitting model parameters.
In short, we recommend the hidden states as the general features in FADS-ICL in most scenarios. 

\subsection{The Role of Demonstrations}
\label{roleofdemons}

In this section, we investigate the role of demonstrations in FADS-ICL, including the impact of whether there is a demonstration and the influence of different numbers of demonstrations. 

\paragraph{Setup} 
We investigate four settings about demonstrations: (1) {\it w/o demon} means that no demonstration is prepended to each sample when extracting the general features. (2) {\it |demon|=1} is the common setting before, {\it i.e.}, one labeled sample per class is chosen as a demonstration. 
(3){\it |demon|=k} denotes that k labeled samples per class are chosen. We only configure this setting when training shots are not less than $2*k$ to ensure that there are enough residual labeled samples left for the supervision of the modulator. (4) {\it |demon|=most} means that as many labeled samples as possible are taken as demonstrations up to the context length.


\paragraph{Result}
\autoref{fig:Ablation}(c) shows that when fully discarding demonstrations, the performance of FADS-ICL (w/o demon) drops dramatically compared with FADS-ICL ($|demon|=1$), verifying the importance of demonstrations. 
Besides, when increasing demonstrations to the maximum, it will improve a little further, but not as significantly as before.

 \begin{figure}[t!]
\centering
 \includegraphics[width=0.48\textwidth]{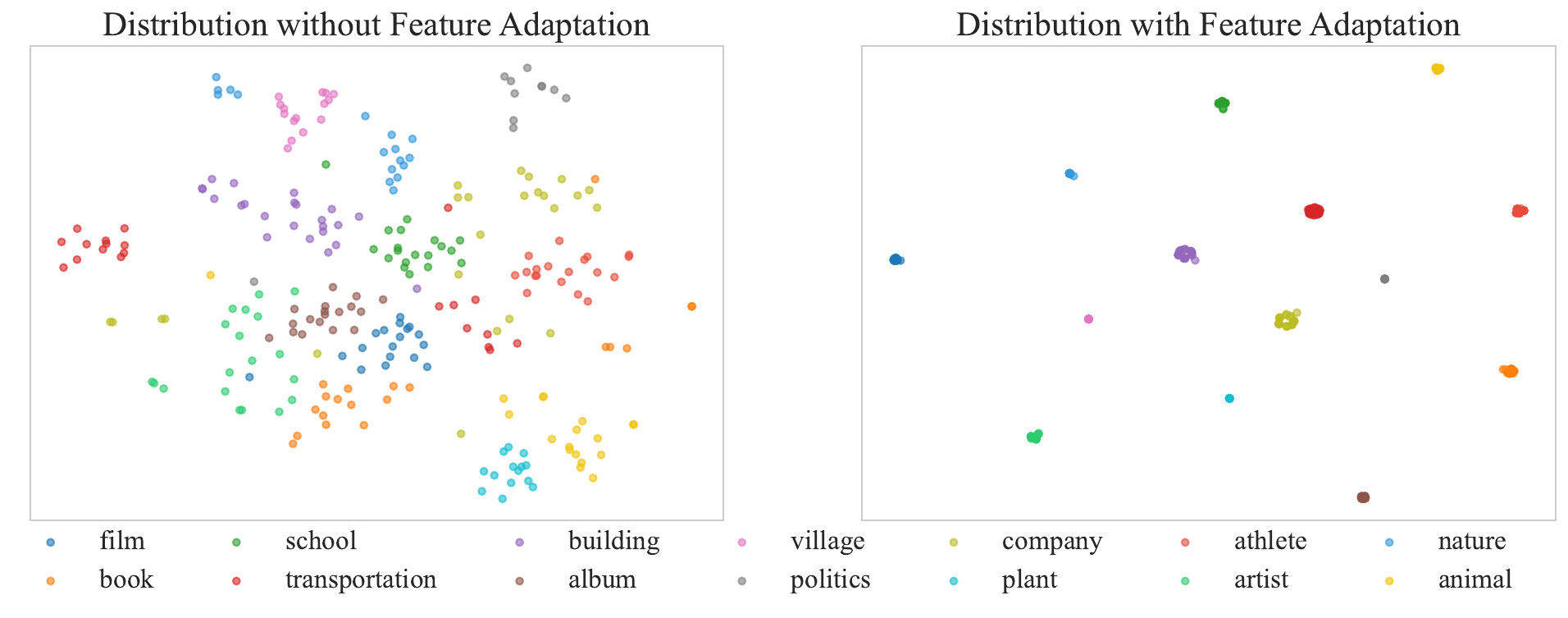}
\caption{Visualization for feature adaptation. }
\label{fig:visualization}
\end{figure}

\paragraph{Analysis and Recommendation}
Demonstrations serve as the context, which can help the LLM better understand the downstream task and partly regularize general features toward the specific task.
From the perspective of tokens, the LLM is designed for language modeling at first, and thus there are many reasonable tokens as candidates, with some task-related and some task-irrelevant. 
Demonstrations can partly filter task-irrelevant tokens and regularize task-related tokens, which is shown in \autoref{APP:demon}. 
Equally for features, the role of demonstrations is feature regularization toward the specific task, which is quite beneficial to further refinement of features in FADS-ICL. 
Finally, we recommend that you take $|demon|=1$ for simplicity, and try to fill the context with demonstrations when sufficient samples are available. 

\subsection{Visualization for Feature Adaptation}
\label{visualization}

We conduct visualization of feature distribution without/with feature adaptation on the DBPedia dataset.
From \autoref{fig:visualization}, we can see that the general features from the LLM maintain a certain degree of discrimination for downstream tasks, but the discrimination of features refined to specific tasks is much higher than that of general features. Therefore, it is essential to refine general features in the LLM to task-specific features to perform the downstream task, that is feature adaptation.

\section{Related Works}

There are many works contributing to improving the performance of ICL on downstream tasks. 
First, most related works focus on designing the best prefix for each test sample, including demonstration selection, corresponding order, and task-specific templates. 
Specifically, \citet{DBLP:conf/acl-deelio/LiuSZDCC22} retrieves demonstration sets based on semantic similarity with an off-shelf sentence encoder, while \citet{DBLP:conf/naacl/RubinHB22,DBLP:journals/corr/abs-2307-07164} train a prompt retriever with weak supervised learning. 
Some works \citep{DBLP:conf/acl/LuBM0S22,DBLP:conf/acl/0003WYK23} explore the order of demonstrations, while other works \citep{ DBLP:conf/acl/SorensenRRSRDKF22, DBLP:conf/eacl/PrasadHZB23} aim to search for the most suitable templates for specific tasks. 
These works do not conflict with FADS-ICL, and you can expect better results when these methods are applied to our framework. 
Second, a few works further finetune the LLM based on the input-output format of ICL with a large set of tasks, to adapt to ICL during inference, {\it e.g.} MetaICL \citep{DBLP:conf/naacl/MinLZH22}, ICT \citep{DBLP:conf/acl/ChenZZK022}. This type of approach has not become a mainstream application paradigm due to the high cost of fine-tuning LLM. 
Third, some works \citep{DBLP:conf/emnlp/ShiMGZ22,DBLP:conf/iclr/Xu0MLS023} have explored exploiting beyond-context samples, similar to KNN-LM \citep{DBLP:conf/iclr/KhandelwalLJZL20}. 
However, they ignore feature refinement for specific tasks and instead directly use features designed for language modeling in LLM, while FADS-ICL implements feature adaptation, resulting in better performance on downstream tasks.

\section{Conclusion}

This paper proposes FADS-ICL, which can leverage task-adaptive features to promote inference on the downstream task with the supervision of beyond-context samples. It implements data scalability by extracting general features using the LLM with ICL input form one by one, while a lightweight task-specific modulator is introduced to complete feature refinement and final prediction, achieving feature adaptation.
FADS-ICL consistently outperforms previous state-of-the-art by a significant margin under all data and LLM scale settings, verifying its effectiveness and superiority. 
Detailed analytical experiments on key settings provide more in-depth insights and recommended configurations in FADS-ICL. 

\section*{Limitation}
The main limitation of FADS-ICL is that it can be easily adapted to any downstream classification tasks, but is difficult to apply to natural language generation tasks. As usually there are only a few or a dozen categories in classification tasks, it is easy to collate several annotated samples for each class. However, natural language generation tasks can be regarded as classification tasks on the entire vocab with tens of thousands of tokens, thus the cost of data annotation becomes unacceptably heavy. 
Besides, the number of parameters in the modulator is also related to the number of categories and its increase cannot be underestimated either. 

\section*{Acknowledgements}
We would like to thank all the reviewers for their valuable suggestions, which significantly improved this paper.
This research is supported by  National Science Fund for Excellent Young Scholars under Grant 62222212 and the General Program of National Natural Science Foundation of China under Grant 62376033.

\bibliography{anthology,custom}
\bibliographystyle{acl_natbib}

\appendix




\section{Performance Across LLM Scales}
\label{APP:Performance Across LLM Scales}

Here, We provide detailed experimental results across LLM scales in \autoref{table:mainresultsacrossLM}. 

\begin{table*}[t]
\centering
\small
\resizebox{2\columnwidth}{!}{
\begin{tabular}{{p{0.6cm}l|cccccccccc|c}}
\toprule
\multicolumn{2}{c|}{\textbf{Models\&Methods}}&\textbf{SST2}&\textbf{SUBJ}&\textbf{MPQA}&\textbf{AGNews}&\textbf{CB}&\textbf{CR}&\textbf{DBPedia}&\textbf{MR}&\textbf{RTE}&\textbf{TREC}&\textbf{AVG}\\
\midrule
\multicolumn{2}{c|}{\small\textbf{BERT Large FT}}&88.3$_{\pm\textrm{1.4}}$&90.7$_{\pm\textrm{0.6}}$&74.5$_{\pm\textrm{5.0}}$&88.0$_{\pm\textrm{1.0}}$&78.6$_{\pm\textrm{3.6}}$&88.0$_{\pm\textrm{2.6}}$&95.1$_{\pm\textrm{1.8}}$&83.0$_{\pm\textrm{3.1}}$&58.1$_{\pm\textrm{1.5}}$&78.8$_{\pm\textrm{5.3}}$&82.3$_{\pm\textrm{2.6}}$\\
\multicolumn{2}{c|}{\small\textbf{GPT Large FT}}&90.7$_{\pm\textrm{1.3}}$&86.1$_{\pm\textrm{1.7}}$&87.6$_{\pm\textrm{0.9}}$&88.3$_{\pm\textrm{1.5}}$&70.0$_{\pm\textrm{2.0}}$&86.7$_{\pm\textrm{13.2}}$&96.5$_{\pm\textrm{1.2}}$&86.2$_{\pm\textrm{1.0}}$&55.4$_{\pm\textrm{3.8}}$&71.2$_{\pm\textrm{2.2}}$&81.9$_{\pm\textrm{2.9}}$\\
\midrule
\multirow{5}*{\textbf{0.8B}}&\small\textbf{ICL}&63.4$_{\pm\textrm7.3}$&58.9$_{\pm\textrm8.7}$&70.5$_{\pm\textrm5.2}$&60.7$_{\pm\textrm12.2}$&37.1$_{\pm\textrm14.5}$&83.3$_{\pm\textrm13.7}$&63.4$_{\pm\textrm6.0}$&77.0$_{\pm\textrm15.7}$&53.6$_{\pm\textrm3.1}$&54.2$_{\pm\textrm8.6}$&62.2$_{\pm\textrm{9.5}}$\\
&\small\textbf{kNN-prompt}&88.0$_{\pm\textrm1.2}$&73.0$_{\pm\textrm7.8}$&76.6$_{\pm\textrm3.0}$&51.4$_{\pm\textrm22.4}$&41.4$_{\pm\textrm0.8}$&86.0$_{\pm\textrm2.5}$&68.0$_{\pm\textrm4.0}$&85.0$_{\pm\textrm3.0}$&52.3$_{\pm\textrm0.0}$&47.7$_{\pm\textrm7.1}$&66.9$_{\pm\textrm{5.2}}$\\
&\small\textbf{kNN-prompting}&83.2$_{\pm\textrm2.8}$&79.8$_{\pm\textrm1.8}$&72.3$_{\pm\textrm7.4}$&86.9$_{\pm\textrm1.8}$&67.5$_{\pm\textrm7.1}$&86.6$_{\pm\textrm2.6}$&95.5$_{\pm\textrm0.7}$&82.4$_{\pm\textrm2.9}$&49.5$_{\pm\textrm1.8}$&75.3$_{\pm\textrm3.8}$&77.9$_{\pm\textrm{3.3}}$\\
&\small\textbf{FADS-ICL}&87.7$_{\pm\textrm2.4}$&90.1$_{\pm\textrm1.9}$&81.7$_{\pm\textrm1.0}$&88.1$_{\pm\textrm1.4}$&80.0$_{\pm\textrm2.3}$&88.7$_{\pm\textrm1.6}$&97.7$_{\pm\textrm0.6}$&85.0$_{\pm\textrm2.2}$&59.4$_{\pm\textrm1.9}$&90.3$_{\pm\textrm1.5}$&\textbf{84.9$_{\pm\textrm{1.7}}$}\\
\midrule
\multirow{5}*{\textbf{1.5B}}&\small\textbf{ICL}&81.3$_{\pm\textrm{5.4}}$&64.1$_{\pm\textrm{11.3}}$&75.2$_{\pm\textrm{8.8}}$&78.2$_{\pm\textrm{6.7}}$&57.9$_{\pm\textrm{9.7}}$&66.2$_{\pm\textrm{16.7}}$&82.0$_{\pm\textrm{2.1}}$&72.2$_{\pm\textrm{13.9}}$&53.0$_{\pm\textrm{1.7}}$&54.8$_{\pm\textrm{4.2}}$&68.5$_{\pm\textrm{8.1}}$\\
&\small\textbf{kNN-prompt}&86.1$_{\pm\textrm3.9}$&87.5$_{\pm\textrm3.3}$&74.2$_{\pm\textrm4.0}$&84.0$_{\pm\textrm5.3}$&42.5$_{\pm\textrm1.5}$&87.0$_{\pm\textrm3.5}$&86.4$_{\pm\textrm2.2}$&87.3$_{\pm\textrm2.1}$&52.3$_{\pm\textrm0.0}$&55.3$_{\pm\textrm4.1}$&74.3$_{\pm\textrm3.0}$\\
&\small\textbf{kNN-prompting}&86.0$_{\pm\textrm2.8}$&83.9$_{\pm\textrm2.2}$&82.1$_{\pm\textrm1.6}$&87.8$_{\pm\textrm1.5}$&65.0$_{\pm\textrm3.0}$&88.8$_{\pm\textrm2.2}$&96.8$_{\pm\textrm1.0}$&86.2$_{\pm\textrm0.8}$&51.4$_{\pm\textrm1.8}$&74.0$_{\pm\textrm3.2}$&80.2$_{\pm\textrm2.0}$\\
&\small\textbf{FADS-ICL}&88.9$_{\pm\textrm0.8}$&92.1$_{\pm\textrm0.9}$&84.8$_{\pm\textrm1.2}$&88.4$_{\pm\textrm1.1}$&83.2$_{\pm\textrm4.5}$&90.5$_{\pm\textrm1.9}$&98.6$_{\pm\textrm0.7}$&86.1$_{\pm\textrm1.0}$&57.8$_{\pm\textrm4.1}$&90.5$_{\pm\textrm1.6}$&\textbf{86.1$_{\pm\textrm1.8}$}\\
\midrule
\multirow{5}*{\textbf{7B}}&\small\textbf{ICL}&93.3$_{\pm\textrm0.7}$&79.6$_{\pm\textrm14.5}$&86.8$_{\pm\textrm1.4}$&86.2$_{\pm\textrm0.9}$&74.6$_{\pm\textrm11.3}$&89.8$_{\pm\textrm2.4}$&81.2$_{\pm\textrm1.5}$&93.3$_{\pm\textrm1.2}$&74.1$_{\pm\textrm1.6}$&83.9$_{\pm\textrm3.9}$&84.3$_{\pm\textrm{3.9}}$\\
&\small\textbf{kNN-prompt}&93.3$_{\pm\textrm0.8}$&82.8$_{\pm\textrm15.3}$&83.6$_{\pm\textrm1.8}$&86.4$_{\pm\textrm2.2}$&80.4$_{\pm\textrm1.8}$&92.3$_{\pm\textrm1.7}$&91.6$_{\pm\textrm1.1}$&92.7$_{\pm\textrm0.3}$&69.8$_{\pm\textrm3.5}$&81.1$_{\pm\textrm6.0}$&85.4$_{\pm\textrm{3.4}}$\\
&\small\textbf{kNN-prompting}&92.8$_{\pm\textrm0.8}$&93.4$_{\pm\textrm1.4}$&84.6$_{\pm\textrm1.3}$&87.6$_{\pm\textrm1.4}$&84.3$_{\pm\textrm2.3}$&91.9$_{\pm\textrm2.1}$&99.0$_{\pm\textrm0.3}$&91.8$_{\pm\textrm0.4}$&70.4$_{\pm\textrm4.7}$&83.4$_{\pm\textrm3.8}$&87.9$_{\pm\textrm{1.8}}$\\
&\small\textbf{FADS-ICL}&92.4$_{\pm\textrm1.0}$&95.6$_{\pm\textrm0.3}$&85.8$_{\pm\textrm1.7}$&89.7$_{\pm\textrm0.7}$&95.0$_{\pm\textrm2.3}$&93.2$_{\pm\textrm1.1}$&98.9$_{\pm\textrm0.4}$&92.4$_{\pm\textrm0.6}$&74.7$_{\pm\textrm0.9}$&93.6$_{\pm\textrm1.8}$&\textbf{91.1$_{\pm\textrm{1.1}}$}\\
\midrule
\multirow{5}*{\textbf{7B$^2$}}&\small\textbf{ICL}&93.6$_{\pm\textrm1.5}$&90.2$_{\pm\textrm1.5}$&87.0$_{\pm\textrm1.8}$&85.5$_{\pm\textrm1.6}$&70.0$_{\pm\textrm7.0}$&90.4$_{\pm\textrm2.4}$&95.7$_{\pm\textrm0.9}$&92.9$_{\pm\textrm1.0}$&74.0$_{\pm\textrm2.1}$&85.5$_{\pm\textrm2.0}$&86.5$_{\pm\textrm{2.2}}$\\
&\small\textbf{kNN-prompt}&94.5$_{\pm\textrm1.0}$&79.5$_{\pm\textrm18.1}$&81.7$_{\pm\textrm3.4}$&85.9$_{\pm\textrm1.7}$&80.0$_{\pm\textrm4.6}$&93.4$_{\pm\textrm1.0}$&98.3$_{\pm\textrm0.5}$&93.2$_{\pm\textrm1.1}$&68.4$_{\pm\textrm4.6}$&84.1$_{\pm\textrm4.9}$&85.9$_{\pm\textrm{4.1}}$\\
&\small\textbf{kNN-prompting}&92.8$_{\pm\textrm1.6}$&95.5$_{\pm\textrm0.6}$&81.7$_{\pm\textrm2.1}$&88.3$_{\pm\textrm1.7}$&77.1$_{\pm\textrm4.6}$&93.7$_{\pm\textrm0.7}$&99.0$_{\pm\textrm0.2}$&92.2$_{\pm\textrm0.6}$&69.1$_{\pm\textrm3.0}$&85.1$_{\pm\textrm2.4}$&86.5$_{\pm\textrm{1.8}}$\\
&\small\textbf{FADS-ICL}&92.2$_{\pm\textrm1.0}$&95.2$_{\pm\textrm0.8}$&87.3$_{\pm\textrm2.1}$&89.1$_{\pm\textrm0.7}$&90.7$_{\pm\textrm3.2}$&93.4$_{\pm\textrm1.2}$&99.1$_{\pm\textrm0.4}$&92.2$_{\pm\textrm0.7}$&75.8$_{\pm\textrm3.1}$&93.0$_{\pm\textrm1.4}$&\textbf{90.8$_{\pm\textrm{1.5}}$}\\
\midrule
\multirow{5}*{\textbf{13B}}&\small\textbf{ICL}&94.9$_{\pm\textrm0.9}$&88.6$_{\pm\textrm2.8}$&85.5$_{\pm\textrm2.0}$&86.2$_{\pm\textrm1.9}$&62.5$_{\pm\textrm6.8}$&87.8$_{\pm\textrm2.7}$&81.5$_{\pm\textrm1.0}$&93.8$_{\pm\textrm0.3}$&71.2$_{\pm\textrm5.1}$&83.4$_{\pm\textrm2.6}$&83.5$_{\pm\textrm{2.6}}$\\
&\small\textbf{kNN-prompt}&93.9$_{\pm\textrm1.0}$&88.3$_{\pm\textrm6.1}$&84.3$_{\pm\textrm2.2}$&89.9$_{\pm\textrm1.0}$&80.0$_{\pm\textrm3.4}$&91.5$_{\pm\textrm0.6}$&95.1$_{\pm\textrm2.9}$&92.7$_{\pm\textrm0.9}$&68.3$_{\pm\textrm3.8}$&75.8$_{\pm\textrm8.3}$
&86.0$_{\pm\textrm{3.0}}$\\
&\small\textbf{kNN-prompting}&93.5$_{\pm\textrm1.1}$&92.2$_{\pm\textrm1.7}$&84.4$_{\pm\textrm0.7}$&88.4$_{\pm\textrm1.1}$&88.9$_{\pm\textrm4.3}$&92.3$_{\pm\textrm1.7}$&99.1$_{\pm\textrm0.3}$&93.1$_{\pm\textrm0.6}$&68.4$_{\pm\textrm1.7}$&84.8$_{\pm\textrm2.0}$
&88.5$_{\pm\textrm{1.5}}$\\
&\small\textbf{FADS-ICL}&93.1$_{\pm\textrm0.3}$&95.7$_{\pm\textrm1.1}$&85.9$_{\pm\textrm1.1}$&89.2$_{\pm\textrm0.4}$&97.1$_{\pm\textrm1.0}$&92.3$_{\pm\textrm0.8}$&99.0$_{\pm\textrm0.3}$&93.4$_{\pm\textrm0.9}$&75.0$_{\pm\textrm3.2}$&93.8$_{\pm\textrm1.4}$&\textbf{91.5$_{\pm\textrm{1.1}}$}\\
\midrule
\multirow{5}*{\textbf{13B$^2$}}&\small\textbf{ICL}&94.9$_{\pm\textrm0.9}$&88.6$_{\pm\textrm2.8}$&85.5$_{\pm\textrm2.0}$&86.2$_{\pm\textrm1.9}$&62.5$_{\pm\textrm6.8}$&87.8$_{\pm\textrm2.7}$&81.5$_{\pm\textrm1.0}$&93.8$_{\pm\textrm0.3}$&71.2$_{\pm\textrm5.1}$&83.4$_{\pm\textrm2.6}$&88.1$_{\pm\textrm{2.1}}$\\
&\small\textbf{kNN-prompt}&93.1$_{\pm\textrm1.0}$&81.0$_{\pm\textrm15.9}$&85.5$_{\pm\textrm1.9}$&87.9$_{\pm\textrm1.4}$&86.1$_{\pm\textrm2.3}$&94.1$_{\pm\textrm0.9}$&97.3$_{\pm\textrm1.3}$&92.7$_{\pm\textrm0.8}$&74.8$_{\pm\textrm4.3}$&77.4$_{\pm\textrm6.3}$
&87.0$_{\pm\textrm{3.6}}$\\
&\small\textbf{kNN-prompting}&94.6$_{\pm\textrm0.9}$&94.6$_{\pm\textrm1.2}$&84.5$_{\pm\textrm1.6}$&87.1$_{\pm\textrm2.2}$&91.4$_{\pm\textrm1.5}$&92.2$_{\pm\textrm2.5}$&98.8$_{\pm\textrm0.2}$&92.6$_{\pm\textrm0.9}$&72.8$_{\pm\textrm3.7}$&86.7$_{\pm\textrm2.3}$
&89.5$_{\pm\textrm{1.7}}$\\
&\small\textbf{FADS-ICL}&93.1$_{\pm\textrm1.0}$&96.2$_{\pm\textrm0.8}$&88.0$_{\pm\textrm0.8}$&89.6$_{\pm\textrm0.7}$&96.1$_{\pm\textrm2.0}$&93.8$_{\pm\textrm0.9}$&98.8$_{\pm\textrm0.3}$&93.4$_{\pm\textrm0.3}$&81.1$_{\pm\textrm3.0}$&95.0$_{\pm\textrm0.8}$&\textbf{92.5$_{\pm\textrm{1.1}}$}\\
\midrule
\multirow{5}*{\textbf{30B}}&\small\textbf{ICL}&94.0$_{\pm\textrm0.3}$&91.2$_{\pm\textrm2.5}$&87.6$_{\pm\textrm1.1}$&87.1$_{\pm\textrm1.8}$&88.2$_{\pm\textrm3.7}$&89.1$_{\pm\textrm2.0}$&81.2$_{\pm\textrm0.5}$&94.7$_{\pm\textrm0.7}$&77.2$_{\pm\textrm3.0}$&84.6$_{\pm\textrm1.7}$&87.5$_{\pm\textrm{1.7}}$\\
&\small\textbf{kNN-prompt}&93.9$_{\pm\textrm1.0}$&89.6$_{\pm\textrm4.4}$&84.3$_{\pm\textrm2.2}$&89.9$_{\pm\textrm1.0}$&80.0$_{\pm\textrm3.4}$&91.5$_{\pm\textrm0.6}$&95.1$_{\pm\textrm2.9}$&92.7$_{\pm\textrm0.9}$&68.3$_{\pm\textrm3.8}$&75.8$_{\pm\textrm8.3}$&86.1$_{\pm\textrm{2.9}}$\\
&\small\textbf{kNN-prompting}&92.8$_{\pm\textrm0.7}$&93.6$_{\pm\textrm1.3}$&84.3$_{\pm\textrm1.6}$&89.0$_{\pm\textrm1.6}$&88.9$_{\pm\textrm2.3}$&92.6$_{\pm\textrm0.8}$&98.8$_{\pm\textrm0.3}$&92.7$_{\pm\textrm0.4}$&74.8$_{\pm\textrm2.3}$&86.6$_{\pm\textrm2.1}$&89.4$_{\pm\textrm{1.3}}$\\
&\small\textbf{FADS-ICL}&92.7$_{\pm\textrm0.6}$&96.4$_{\pm\textrm0.7}$&87.3$_{\pm\textrm1.6}$&90.5$_{\pm\textrm0.7}$&94.6$_{\pm\textrm0.0}$&91.9$_{\pm\textrm1.7}$&98.9$_{\pm\textrm0.3}$&93.8$_{\pm\textrm1.6}$&81.8$_{\pm\textrm2.0}$&95.2$_{\pm\textrm0.8}$&\textbf{92.3$_{\pm\textrm{1.0}}$}\\
\midrule
\multirow{5}*{\textbf{70B$^2$}}&\small\textbf{ICL}&93.2$_{\pm\textrm1.6}$&96.5$_{\pm\textrm1.1}$&89.1$_{\pm\textrm1.3}$&88.8$_{\pm\textrm1.3}$&95.4$_{\pm\textrm2.7}$&92.3$_{\pm\textrm1.9}$&94.1$_{\pm\textrm1.6}$&93.9$_{\pm\textrm0.4}$&85.5$_{\pm\textrm0.3}$&88.2$_{\pm\textrm3.9}$&91.7$_{\pm\textrm{1.6}}$\\
&\small\textbf{kNN-prompt}&92.0$_{\pm\textrm1.4}$&86.0$_{\pm\textrm5.2}$&84.3$_{\pm\textrm1.6}$&87.5$_{\pm\textrm1.6}$&87.9$_{\pm\textrm3.2}$&92.5$_{\pm\textrm1.4}$&89.5$_{\pm\textrm3.1}$&92.4$_{\pm\textrm1.3}$&79.4$_{\pm\textrm2.7}$&87.6$_{\pm\textrm2.6}$&87.9$_{\pm\textrm{2.4}}$\\
&\small\textbf{kNN-prompting}&91.9$_{\pm\textrm2.2}$&91.3$_{\pm\textrm3.6}$&84.1$_{\pm\textrm1.5}$&87.7$_{\pm\textrm1.4}$&94.3$_{\pm\textrm0.8}$&92.0$_{\pm\textrm1.4}$&98.8$_{\pm\textrm0.4}$&91.6$_{\pm\textrm1.3}$&80.2$_{\pm\textrm2.9}$&90.1$_{\pm\textrm2.6}$&90.2$_{\pm\textrm{1.8}}$\\
&\small\textbf{FADS-ICL}&92.6$_{\pm\textrm1.1}$&96.3$_{\pm\textrm0.7}$&87.7$_{\pm\textrm2.1}$&90.2$_{\pm\textrm1.0}$&98.2$_{\pm\textrm1.3}$&93.3$_{\pm\textrm1.6}$&98.9$_{\pm\textrm0.3}$&93.0$_{\pm\textrm0.5}$&83.6$_{\pm\textrm1.7}$&95.2$_{\pm\textrm1.5}$&\textbf{92.9$_{\pm\textrm{1.2}}$}\\
\bottomrule
\end{tabular}}
\caption{Results across LLM scales with 128-shots. Here, 0.8B and 1.5B denote the GPT-2 series; 7B, 13B, and 30B denote the Llama-1 series; 7B$^2$, 13B$^2$ and 70B$^2$ denotes the Llama-2 series. }
\label{table:mainresultsacrossLM}
\end{table*}

\section{Comparison to Baselines based on Demonstration Selection}
\label{APP:comparisontoselection}

In this subsection, We compare FADS-ICL with KATE \citep{DBLP:conf/acl-deelio/LiuSZDCC22} based on demonstration selection via similarity measure. Specifically, we consider multiple similarity measurement methods and compare experimental results when available samples are beyond the context. The full result is presented in \autoref{table:comparisontoselection}. We can see that KATEs based on all similarity measurement methods have provided a slight performance improvement, which is completely incomparable with FADS-ICL.

\section{The Role of Demonstrations in ICL}
\label{APP:demon}

\begin{figure}[t!]
\centering
\includegraphics[width=0.48\textwidth]{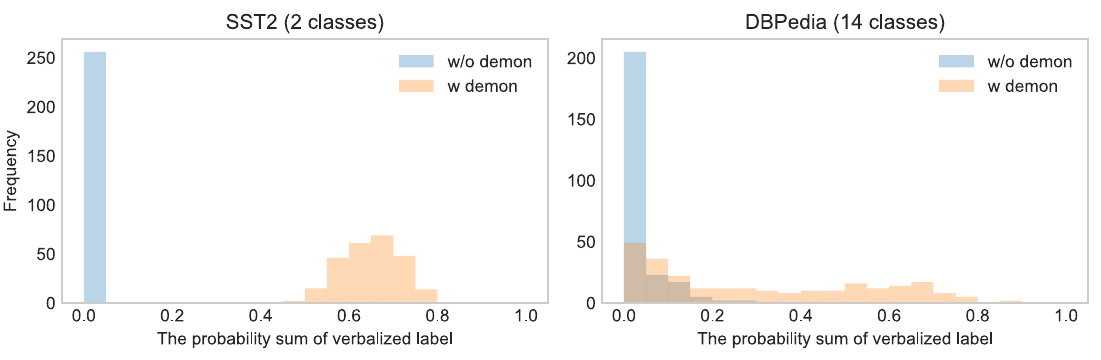}
\caption{The impact of the demonstrations on the probability distribution of verbalized labels in vanilla ICL. }
\label{fig:SFC}
\end{figure}

\autoref{fig:SFC} presents the distributions of the probability sum of task-specific labels without/with demonstrations in vanilla ICL. It shows that the probability sum of task-specific labels is quite low without demonstrations, and demonstrations improve it significantly. This shows that demonstrations can partly filter task-irrelevant tokens and regularize task-related tokens. 

\section{Black-box LLMs}
\label{APP:blackboxllm}

In the main experiment, the effectiveness of the method has been verified based on various open-source LLMs (GPT-2, Llama-1, and Llama-2 series). For those black-box LLMs mainly used for chat (i.e. text generation), our method is indeed not applicable because the features that we need in the LLM cannot be obtained through the API. Fortunately, most AI companies in the black-box LLM business have a dedicated business branch to provide feature output for users, such as text-embedding-ada-002\footnote{https://openai.com/blog/new-and-improved-embedding-model} released by OpenAI, text-embedding-v2\footnote{https://help.aliyun.com/zh/dashscope/developer-reference/generic-text-vector} released by Alibaba Cloud, etc. We experiment on OpenAI 's text-embedding-ada-002. Since this black-box LLM does not provide the text generation function, vanilla ICL is unavailable. We give the previous SOTA method, kNN-prompting, as a comparison and the experimental result is present in \autoref{table:blackboxllm}. It shows that the proposed method provides significant improvements compared to kNN-prompting, which is consistent with the conclusion in the main experiment.
\begin{table}[!t]
\centering
\resizebox{\columnwidth}{!}{
\begin{tabular}{{lccc}}
\toprule
\textbf{Methods}&\textbf{8-shots}&\textbf{32-shots}&\textbf{128-shots}\\
\midrule
\textbf{kNN-prompting}&61.4&68.4&72.9\\
\textbf{FADS-ICL}&\textbf{73.0}&\textbf{80.8}&\textbf{87.0}\\
\bottomrule
\end{tabular}}
\caption{The performance of FADS-ICL on the black-box LLM. The experiment is conducted on OpenAI's text-embedding-ada-002 model. } 
\label{table:blackboxllm}
\end{table}

\begin{table}[!t]
\centering
\resizebox{\columnwidth}{!}{
\begin{tabular}{{lccc}}
\toprule
\textbf{Methods}&\textbf{32-shots}&\textbf{64-shots}&\textbf{128-shots}\\
\midrule
\textbf{kNN-prompting}&76.4&78.7&79.7\\
\textbf{Fine Tuning (0.8B)}&74.7&75.8&81.9\\
\textbf{LoRA (1.5M)}&71.4&73.0&77.1\\
\textbf{FADS-ICL (2.6K)}&\textbf{81.7}&\textbf{83.7}&\textbf{84.9}\\
\bottomrule
\end{tabular}}
\caption{The comparison of FADS-ICL, fine-tuning and parameter-efficient fine-tuning. The experiment is conducted on GPT2-Large and inside the parentheses are the number of parameters involved in training. } 
\label{table:PEFT}
\end{table}

\begin{table*}[h]
\centering
\resizebox{2\columnwidth}{!}{
\begin{tabular}{{ll|cccccccccc|c}}
\toprule
\multicolumn{2}{c|}{\textbf{Setting \& Methods}}&\textbf{SST2}&\textbf{SUBJ}&\textbf{MPQA}&\textbf{AGNews}&\textbf{CB}&\textbf{CR}&\textbf{DBPedia}&\textbf{MR}&\textbf{RTE}&\textbf{TREC}&\textbf{AVG}\\
\midrule
\multicolumn{2}{c|}{\small\textbf{In-Context Learning}}&81.3$_{\pm\textrm{5.4}}$&64.1$_{\pm\textrm{11.3}}$&75.2$_{\pm\textrm{8.8}}$&72.7$_{\pm\textrm{18.5}}$&60.7$_{\pm\textrm{2.8}}$&66.2$_{\pm\textrm{16.7}}$&83.5$_{\pm\textrm{3.8}}$&72.2$_{\pm\textrm{13.9}}$&53.0$_{\pm\textrm{1.7}}$&54.2$_{\pm\textrm{4.9}}$&68.3\\
\midrule
\multirow{5}*{$m=32$}&\small\textbf{BM25}&68.4$_{\pm\textrm{3.7}}$&63.6$_{\pm\textrm{3.0}}$&75.2$_{\pm\textrm{8.8}}$&69.0$_{\pm\textrm{3.6}}$&65.0$_{\pm\textrm{6.5}}$&55.2$_{\pm\textrm{1.0}}$&80.7$_{\pm\textrm{1.3}}$&59.4$_{\pm\textrm{1.4}}$&53.0$_{\pm\textrm{2.3}}$&65.5$_{\pm\textrm{2.8}}$&65.5\\
&\small\textbf{SBERT}&71.0$_{\pm\textrm{4.9}}$&67.5$_{\pm\textrm{1.6}}$&75.2$_{\pm\textrm{8.8}}$&82.3$_{\pm\textrm{2.1}}$&62.9$_{\pm\textrm{2.9}}$&57.8$_{\pm\textrm{1.9}}$&83.8$_{\pm\textrm{1.2}}$&57.7$_{\pm\textrm{4.3}}$&51.2$_{\pm\textrm{3.1}}$&58.4$_{\pm\textrm{6.7}}$&66.8\\
&\small\textbf{SimCSE}&68.1$_{\pm\textrm{4.5}}$&69.8$_{\pm\textrm{3.1}}$&75.2$_{\pm\textrm{8.8}}$&80.7$_{\pm\textrm{2.9}}$&66.4$_{\pm\textrm{2.3}}$&55.9$_{\pm\textrm{2.8}}$&82.3$_{\pm\textrm{1.9}}$&57.3$_{\pm\textrm{2.4}}$&52.8$_{\pm\textrm{1.7}}$&54.5$_{\pm\textrm{1.8}}$&66.3\\
&\small\textbf{Trans-Encoder}&67.9$_{\pm\textrm{3.8}}$&70.9$_{\pm\textrm{4.1}}$&75.2$_{\pm\textrm{8.8}}$&77.7$_{\pm\textrm{2.5}}$&61.4$_{\pm\textrm{6.0}}$&56.7$_{\pm\textrm{1.7}}$&82.8$_{\pm\textrm{2.0}}$&57.3$_{\pm\textrm{4.3}}$&52.7$_{\pm\textrm{2.0}}$&59.3$_{\pm\textrm{3.9}}$&66.2\\
&\small\textbf{FADS-ICL}&88.0$_{\pm\textrm4.6}$&90.1$_{\pm\textrm2.2}$&81.4$_{\pm\textrm3.0}$&84.9$_{\pm\textrm1.4}$&72.1$_{\pm\textrm2.0}$&92.3$_{\pm\textrm0.7}$&97.9$_{\pm\textrm0.9}$&85.9$_{\pm\textrm2.8}$&55.5$_{\pm\textrm2.9}$&79.9$_{\pm\textrm4.7}$&\textbf{82.8}\\
\midrule
\multirow{5}*{$m=64$}&\small\textbf{BM25}&69.7$_{\pm\textrm{2.9}}$&67.7$_{\pm\textrm{2.5}}$&79.4$_{\pm\textrm{2.0}}$&71.7$_{\pm\textrm{1.9}}$&68.6$_{\pm\textrm{4.1}}$&54.7$_{\pm\textrm{1.1}}$&83.4$_{\pm\textrm{2.0}}$&58.7$_{\pm\textrm{1.5}}$&52.0$_{\pm\textrm{1.6}}$&65.9$_{\pm\textrm{5.6}}$&67.2\\
&\small\textbf{SBERT}&71.8$_{\pm\textrm{3.4}}$&71.6$_{\pm\textrm{3.8}}$&80.2$_{\pm\textrm{1.5}}$&84.6$_{\pm\textrm{2.2}}$&66.8$_{\pm\textrm{6.8}}$&59.8$_{\pm\textrm{1.8}}$&84.6$_{\pm\textrm{0.8}}$&57.6$_{\pm\textrm{1.6}}$&52.8$_{\pm\textrm{3.6}}$&63.7$_{\pm\textrm{4.7}}$&69.4\\
&\small\textbf{SimCSE}&68.4$_{\pm\textrm{4.7}}$&69.7$_{\pm\textrm{2.6}}$&81.6$_{\pm\textrm{0.6}}$&83.1$_{\pm\textrm{2.2}}$&71.4$_{\pm\textrm{4.6}}$&57.9$_{\pm\textrm{2.6}}$&84.8$_{\pm\textrm{2.1}}$&57.3$_{\pm\textrm{2.0}}$&52.3$_{\pm\textrm{3.6}}$&58.1$_{\pm\textrm{1.8}}$&68.5\\
&\small\textbf{Trans-Encoder}&69.3$_{\pm\textrm{4.3}}$&73.0$_{\pm\textrm{1.2}}$&82.0$_{\pm\textrm{2.2}}$&79.3$_{\pm\textrm{2.0}}$&69.3$_{\pm\textrm{3.4}}$&57.9$_{\pm\textrm{2.7}}$&85.6$_{\pm\textrm{1.2}}$&57.9$_{\pm\textrm{1.4}}$&52.1$_{\pm\textrm{4.5}}$&60.1$_{\pm\textrm{2.6}}$&68.7\\
&\small\textbf{FADS-ICL}&88.2$_{\pm\textrm1.8}$&90.5$_{\pm\textrm1.7}$&83.6$_{\pm\textrm1.5}$&87.8$_{\pm\textrm1.0}$&78.6$_{\pm\textrm3.6}$&91.2$_{\pm\textrm0.8}$&98.1$_{\pm\textrm0.7}$&87.0$_{\pm\textrm1.3}$&58.7$_{\pm\textrm3.7}$&84.4$_{\pm\textrm4.0}$&\textbf{84.8}\\
\midrule
\multirow{5}*{$m=128$}&\small\textbf{BM25}&69.1$_{\pm\textrm{0.5}}$&66.8$_{\pm\textrm{2.7}}$&75.2$_{\pm\textrm{6.2}}$&77.5$_{\pm\textrm{1.4}}$&71.4$_{\pm\textrm{0.0}}$&56.4$_{\pm\textrm{1.2}}$&85.5$_{\pm\textrm{1.7}}$&59.8$_{\pm\textrm{2.0}}$&54.5$_{\pm\textrm{1.3}}$&72.3$_{\pm\textrm{4.9}}$&68.9\\
&\small\textbf{SBERT}&71.7$_{\pm\textrm{1.9}}$&71.9$_{\pm\textrm{1.7}}$&79.6$_{\pm\textrm{3.6}}$&85.3$_{\pm\textrm{2.0}}$&69.6$_{\pm\textrm{0.0}}$&58.8$_{\pm\textrm{0.8}}$&87.2$_{\pm\textrm{0.9}}$&60.1$_{\pm\textrm{2.3}}$&53.3$_{\pm\textrm{2.3}}$&64.9$_{\pm\textrm{2.1}}$&70.2\\
&\small\textbf{SimCSE}&70.9$_{\pm\textrm{2.2}}$&71.6$_{\pm\textrm{3.2}}$&81.6$_{\pm\textrm{2.6}}$&84.5$_{\pm\textrm{1.7}}$&73.2$_{\pm\textrm{0.0}}$&58.5$_{\pm\textrm{2.2}}$&87.0$_{\pm\textrm{2.1}}$&58.7$_{\pm\textrm{1.4}}$&53.4$_{\pm\textrm{3.2}}$&59.6$_{\pm\textrm{3.3}}$&69.9\\
&\small\textbf{Trans-Encoder}&69.0$_{\pm\textrm{1.3}}$&75.5$_{\pm\textrm{2.2}}$&82.6$_{\pm\textrm{1.9}}$&82.9$_{\pm\textrm{0.6}}$&73.2$_{\pm\textrm{0.0}}$&56.1$_{\pm\textrm{0.8}}$&87.0$_{\pm\textrm{1.4}}$&57.1$_{\pm\textrm{1.3}}$&52.9$_{\pm\textrm{3.1}}$&63.9$_{\pm\textrm{2.3}}$&70.0\\
&\small\textbf{FADS-ICL}&88.9$_{\pm\textrm0.8}$&92.1$_{\pm\textrm0.9}$&84.8$_{\pm\textrm1.2}$&88.4$_{\pm\textrm1.1}$&83.2$_{\pm\textrm4.5}$&90.5$_{\pm\textrm1.9}$&98.6$_{\pm\textrm0.7}$&86.1$_{\pm\textrm1.0}$&57.8$_{\pm\textrm4.1}$&90.5$_{\pm\textrm1.6}$&\textbf{86.1}\\
\midrule
\multirow{5}*{$m=256$}&\small\textbf{BM25}&72.0$_{\pm\textrm{3.9}}$&72.3$_{\pm\textrm{1.4}}$&78.8$_{\pm\textrm{3.2}}$&77.3$_{\pm\textrm{3.0}}$&71.4$_{\pm\textrm{0.0}}$&57.1$_{\pm\textrm{0.8}}$&88.2$_{\pm\textrm{1.9}}$&58.4$_{\pm\textrm{1.6}}$&53.8$_{\pm\textrm{2.8}}$&76.6$_{\pm\textrm{3.1}}$&70.6\\
&\small\textbf{SBERT}&69.9$_{\pm\textrm{1.8}}$&72.3$_{\pm\textrm{0.8}}$&82.2$_{\pm\textrm{2.5}}$&86.3$_{\pm\textrm{1.2}}$&69.6$_{\pm\textrm{0.0}}$&58.8$_{\pm\textrm{1.5}}$&88.5$_{\pm\textrm{0.9}}$&59.8$_{\pm\textrm{2.6}}$&52.3$_{\pm\textrm{2.4}}$&69.1$_{\pm\textrm{0.8}}$&70.9\\
&\small\textbf{SimCSE}&71.4$_{\pm\textrm{3.7}}$&73.7$_{\pm\textrm{1.6}}$&82.9$_{\pm\textrm{0.8}}$&85.5$_{\pm\textrm{1.4}}$&73.2$_{\pm\textrm{0.0}}$&59.7$_{\pm\textrm{1.2}}$&89.1$_{\pm\textrm{1.7}}$&57.8$_{\pm\textrm{2.2}}$&51.1$_{\pm\textrm{2.6}}$&64.3$_{\pm\textrm{1.5}}$&70.9\\
&\small\textbf{Trans-Encoder}&70.0$_{\pm\textrm{1.0}}$&76.6$_{\pm\textrm{1.4}}$&82.1$_{\pm\textrm{2.0}}$&84.1$_{\pm\textrm{1.2}}$&73.2$_{\pm\textrm{0.0}}$&58.0$_{\pm\textrm{1.0}}$&89.9$_{\pm\textrm{1.3}}$&58.1$_{\pm\textrm{1.2}}$&52.0$_{\pm\textrm{2.4}}$&70.9$_{\pm\textrm{2.2}}$&71.5\\
&\small\textbf{FADS-ICL}&90.5$_{\pm\textrm1.2}$&91.6$_{\pm\textrm1.7}$&84.2$_{\pm\textrm1.9}$&89.8$_{\pm\textrm0.5}$&83.2$_{\pm\textrm4.5}$&89.6$_{\pm\textrm1.7}$&98.7$_{\pm\textrm0.3}$&86.2$_{\pm\textrm1.1}$&59.2$_{\pm\textrm2.0}$&91.0$_{\pm\textrm0.8}$&\textbf{86.4}\\
\bottomrule
\end{tabular}}
\caption{Results for comparison to baselines based on demonstration selection.}
\label{table:comparisontoselection}
\end{table*}

\section{FADS-ICL vs PEFT}

We provide below a comparison of FADS-ICL, fine-tuning, and parameter-efficient fine-tuning (PEFT) under various data settings. Experiments prove that fine-tuning and PEFT are not suitable for low-resource scenarios. 

As can be seen from \autoref{table:PEFT}, in the case of low resources (such as 32, 64-shot), the performance of the fine-tuning method is far inferior to KNN-prompting and FADS-ICL based on in-context learning because training data is far from sufficient to support transfer learning from the original language modeling task to downstream tasks. When the available data grows to 128-shot, the fine-tuning method improves rapidly and becomes competitive, but there is still a certain gap from FADS-ICL. 

In addition, PEFT, such as LoRA, is also not suitable for low-resource scenarios due to their high instability. It can be seen from the overall results that although Lora reduces the training parameters to a certain extent, the average performance is far inferior to the fine-tuning method and FADS-ICL. Careful inspection revealed that due to the small amount of training data, Lora's training was extremely unstable and could easily lead to model collapse. For example, in five random runs on the SST2 dataset, the highest accuracy can reach 93.8 while the lowest is only 64.5 with only different seeds, and training on CR resulted in model collapse with performance approaching random guessing. As pointed out in previous work \citep{DBLP:conf/emnlp/Chen0ML22}, PEFT (including Adapter, Prompt tuning, LoRA, and BitFit) exhibit high instability in low resource settings subject to newly introduced parameter initialization and training data order, determined by random seeds.

Compared with them, FADS-ICL has the following advantages in low-resource scenarios: much better performance, higher stability, and negligible training overhead (within 1s CPU).

\section{Prompt Template}
\label{appendix:template}

The used templates are presented in \autoref{table:icltemplate} (adopted from \citet{lu-etal-2022-fantastically}), which are intuitively designed.

\newcolumntype{L}[1]{>{\raggedright\arraybackslash}p{#1}}
\newcolumntype{C}[1]{>{\centering\arraybackslash}p{#1}}
\newcolumntype{R}[1]{>{\raggedleft\arraybackslash}p{#1}}
\begin{table*}[!h]
\centering
\scriptsize
\begin{tabularx}{2\columnwidth}{l|X|L{2cm}}
\toprule
\textbf{Task}&\textbf{Template}&\textbf{Label Space}\\
\midrule
\multirow[t]{4}{*}{\textbf{SST2}}&Review: contains no wit , only labored gags&negative, positive\\
&Sentiment: negative&\\
&Review: the film is powerful , accessible and funny .&\\
&Sentiment:&\\
\midrule
\multirow[t]{4}{*}{\textbf{SUBJ}}&Input: the script isn't very good ; not even someone as gifted as hoffman ( the actor ) can make it work .&\multirow[t]{4}*{subjective, objective}\\
&Type: subjective&\\
&Input: he must do this in secret so that the parents and school personnel know nothing of his plan .&\\
&Type:&\\
\midrule
\multirow[t]{4}{*}{\textbf{MPQA}}&Review: would not find it at all strange&negative, positive\\
&Sentiment: negative&\\
&Review: as small ( yet acceptable ) as possible&\\
&Sentiment:&\\
\midrule
\multirow[t]{4}{*}{\textbf{AGNews}}&Input: Carlyle Looks Toward Commercial Aerospace (Reuters). "Reuters - Private investment firm Carlyle Group, which has a reputation for making well-timed and occasionally controversial plays in the defense industry, has quietly placed its bets on another part of the market.&world, sports, business, technology\\
&Type: technology&\\
&Input: Superstar Kewell remains centre of attention. Socceroo forward Harry Kewell loosens up by tossing around a ball at Bondi beach yesterday. Photo: Craig Golding. There were half a dozen Socceroos standing on a raised platform in Sydney \#39;s&\\
&Type:&\\
\midrule
\multirow[t]{6}{*}{\textbf{CB}}&Premise: It was a complex language. Not written down but handed down. One might say it was peeled down.&False, True, Neither\\
&Hypothesis: the language was peeled down&\\
&Prediction: False&\\
&Premise: A: so I don't know if I wasn't drug tested based on that or because the man who hired me didn't request the drug test, because I know that my company does drug testing on occasion. B: Right. Well, for instance, does the company you worked for before have the right or do they have the ability to say, hey, we've already drug tested her and she came up negative. A: Well, no, I don't think they can force another company to not drug test me just by saying that I didn't, I mean,&\\
&Hypothesis: they can force another company to not drug test her&\\
&Prediction:&\\
\midrule
\multirow[t]{4}{*}{\textbf{CR}}&Review: it 's not as stylized as a sony or samsung .&negative, positive\\
&Sentiment: negative&\\
&Review: i went out and got the canon today .&\\
&Sentiment:&\\
\midrule
\multirow[t]{4}{*}{\textbf{DBPedia}}&Input: Geoffrey D. Falksen (born July 31 1982) is an American steampunk writer.&company, school,\\
&Type: artist&artist, athlete,\\
&Input: Monster Night is a 2006 film directed by Leslie Allen and Lorenzo Doumani.&politics,\\
&Type:&transportation, building, nature, village, animal, plant, album, film, book\\
\midrule
\multirow[t]{4}{*}{\textbf{MR}}&Review: "you might say tykwer has done all that heaven allows , if you wanted to make as anti-kieslowski a pun as possible . suffice to say its total promise is left slightly unfulfilled ."&negative, positive\\
&Sentiment: negative&\\
&Review: an alternately raucous and sappy ethnic sitcom . . . you'd be wise to send your regrets .&\\
&Sentiment:&\\
\midrule
\multirow[t]{6}{*}{\textbf{RTE}}&Premise: A man is due in court later charged with the murder 26 years ago of a teenager whose case was the first to be featured on BBC One's Crimewatch. Colette Aram, 16, was walking to her boyfriend's house in Keyworth, Nottinghamshire, on 30 October 1983 when she disappeared. Her body was later found in a field close to her home. Paul Stewart Hutchinson, 50, has been charged with murder and is due before Nottingham magistrates later."&false, true\\
&Hypothesis: Paul Stewart Hutchinson is accused of having stabbed a girl.&\\
&Prediction: false&\\
&Premise: For women earning 22,000 a year, the total pay accumulated after six months maternity leave would be just 5,300 in the UK and 5,850 in Ireland. Entitlements in Germany would also be relatively low, at 5,900, along with those in France, Spain and the Netherlands, all at 6,750. At the other end of the scale, pay received after six months leave in Italy would be 9,150 while in Denmark and Norway it would be as much as 11,000.&\\
&Hypothesis: Maternity leave varies in Europe.&\\
&Prediction:&\\
\midrule
\multirow[t]{4}{*}{\textbf{TREC}}&Question: How did serfdom develop in and then leave Russia ?&description, entity, \\
&Type: description&expression, human, \\
&Question: What is Shakespeare 's nickname ?&location, number\\
&Type:&\\
\bottomrule
\end{tabularx}
\caption{Templates for ICL. These are minimum cases with only one demonstration example for illustration.}
\label{table:icltemplate}
\end{table*}

\end{document}